\documentclass[10pt,twocolumn,letterpaper]{article}

\usepackage{iccv}
\usepackage{times}
\usepackage{epsfig}
\usepackage{graphicx}
\usepackage{amsmath}
\usepackage{amssymb}
\usepackage{booktabs}
\usepackage{tikz,pgfplots}
\usepackage{multirow}
\usepackage{array}

\usepackage[breaklinks=true,bookmarks=false]{hyperref}

\iccvfinalcopy 

\def\iccvPaperID{****} 

\ificcvfinal\fi

\begin{document}

\setlength{\abovedisplayskip}{2pt}
\setlength{\belowdisplayskip}{4pt}
\setlength{\abovedisplayshortskip}{2pt}
\setlength{\belowdisplayshortskip}{4pt}

\title{HMD-NeMo: Online 3D Avatar Motion Generation From Sparse Observations}

\author{
Sadegh Aliakbarian
\enspace
\quad
Fatemeh Saleh
\enspace
\quad
David Collier
\enspace
\quad
Pashmina Cameron
\enspace
\quad
Darren Cosker
\\
\\
Microsoft Mixed Reality \& AI Lab, Cambridge, UK
}

\maketitle
\ificcvfinal\thispagestyle{empty}\fi

\begin{abstract}
   Generating both plausible and accurate full body avatar motion is the key to the quality of immersive experiences in mixed reality scenarios. Head-Mounted Devices (HMDs) typically only provide a few input signals, such as head and hands 6-DoF. Recently, different approaches achieved impressive performance in generating full body motion given only head and hands signal. However, to the best of our knowledge, all existing approaches rely on full hand visibility. While this is the case when, e.g., using motion controllers, a considerable proportion of mixed reality experiences do not involve motion controllers and instead rely on egocentric hand tracking. This introduces the challenge of partial hand visibility owing to the restricted field of view of the HMD. In this paper, we propose the first unified approach, HMD-NeMo, that addresses plausible and accurate full body motion generation even when the hands may be only partially visible. HMD-NeMo is a lightweight neural network that predicts the full body motion in an online and real-time fashion. At the heart of HMD-NeMo is the spatio-temporal encoder with novel temporally adaptable mask tokens that encourage plausible motion in the absence of hand observations. We perform extensive analysis of the impact of different components in HMD-NeMo and introduce a new state-of-the-art on AMASS dataset through our evaluation.
\end{abstract}

\section{Introduction}
\label{sec:introduction}
Mixed reality technology opens up new means of communication and interaction between people. With \emph{people} at the heart of this technology, generating faithful and believable avatar motion is key to the quality of immersive experiences. Despite great advances in this area, generating full body avatar motion given HMD signal remains a challenge: in many current solutions, avatars only have upper bodies.

\begin{figure}
    \centering
     \includegraphics[width=0.45\textwidth]{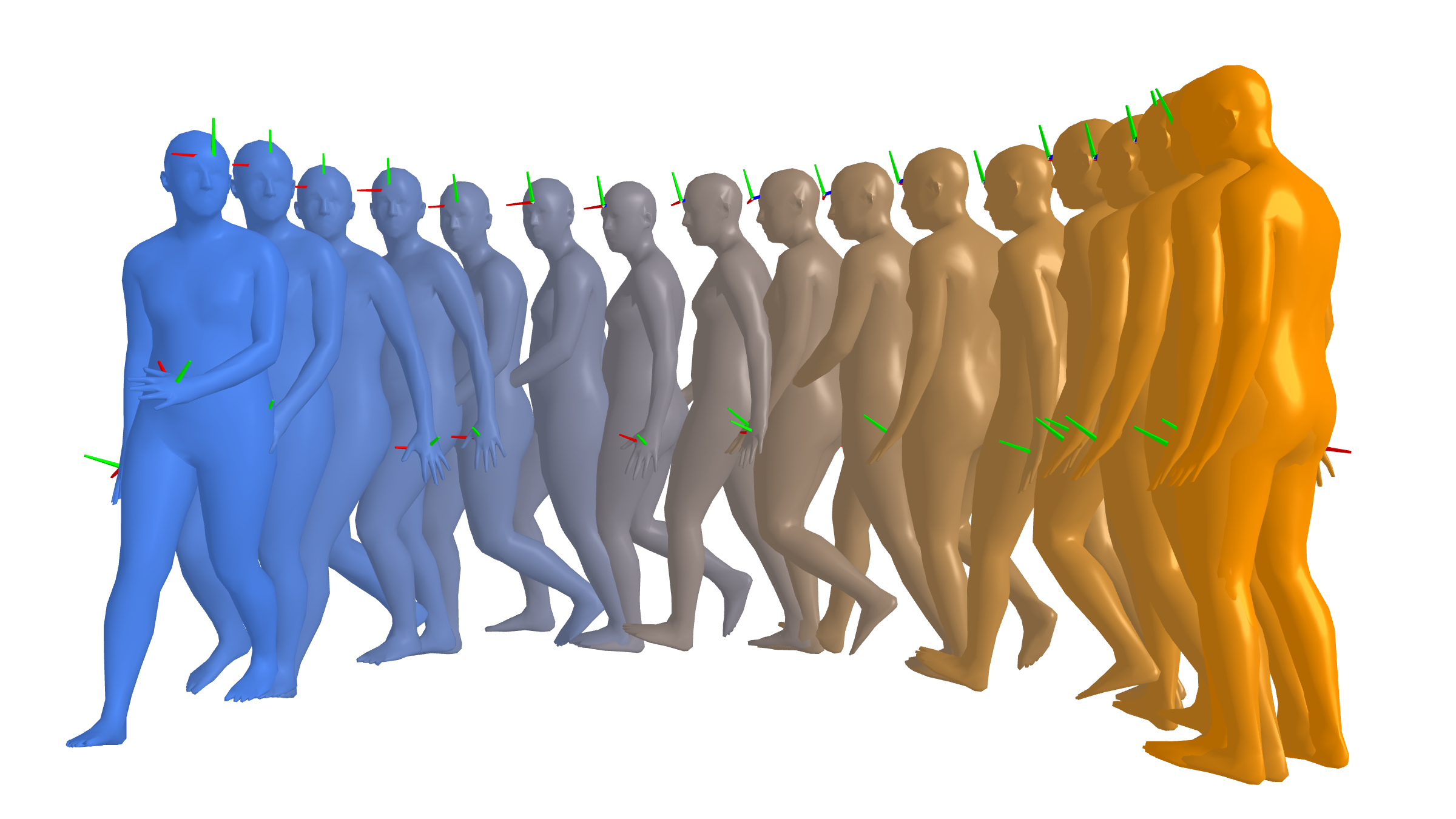}
     \vspace{-10pt}
    \caption{HMD-NeMo generates full body avatar motion given HMD signals, i.e., head and hand 6-DoFs, from hand tracking signal from the HMD as well as hand motion controllers.}
    \label{fig:teaser}
\end{figure}

Prior works attempted to generate full body avatar motion given sparse or partial observations, such as images~\cite{kolotouros2021probabilistic,biggs20203d,zanfir2020weakly}, 2D joints/keypoints~\cite{pavlakos2019expressive,bogo2016keep}, markers~\cite{zhang2021we,loper2014mosh,zanfir2021thundr,ghorbani2021soma}, and IMUs~\cite{von2017sparse,huang2018deep,yi2021transpose,yi2022physical}. While such observations are considered \emph{partial} or \emph{sparse}, they provide much richer input signal compared to a typical HMD's head and hand 6-DoF. More recently, great progress has been made to generate full body motion given only a HMD signal~\cite{ahuja2021coolmoves,yang2021lobstr,ponton2022mmvr,jiang2022avatarposer}, however, they all rely on the availability and full visibility of both hands. While this is the case when, e.g., using motion controllers, many mixed reality experiences do not involve motion controllers and instead rely on hand tracking. This introduces the challenge of partial hand visibility owing to the restricted field of view (FoV) of the HMD sensors.

In this paper, we address this problem via HMD-NeMo (a \textbf{ne}ural \textbf{mo}tion model of human given \textbf{HMD} signal). Within a unified framework, HMD-NeMo generates full body motion in real time, regardless of whether hands are fully or partially observed, or not observed at all. Our approach is built upon recurrent neural networks to efficiently capture temporal information, and a transformer to capture complex relations between different components of the input signal. At the heart of our approach is the TAMT (\textbf{t}emporally \textbf{a}daptable \textbf{m}ask \textbf{t}oken) module, allowing us to handle missing hand observations. 

Our contributions are: (1) The first full body avatar motion generation approach capable of generating accurate and plausible motions with full or partial hand visibility.
This is a step forward for unlocking fully immersive experiences in mixed reality with fewer limitations on the hardware.  (2) Temporally adaptable Mask Tokens (TAMT), a simple yet effective strategy for handling missing hand observations in a temporally coherent way. (3) Extensive experiments and ablations of our method as well as a new state-of-the-art performance on the challenging AMASS dataset.

\begin{figure}
\centering
\scriptsize
\begin{tabular}{@{}c @{ } c @{ } c}
\includegraphics[trim={200px 0 100px 0}, clip, width=0.33\columnwidth]{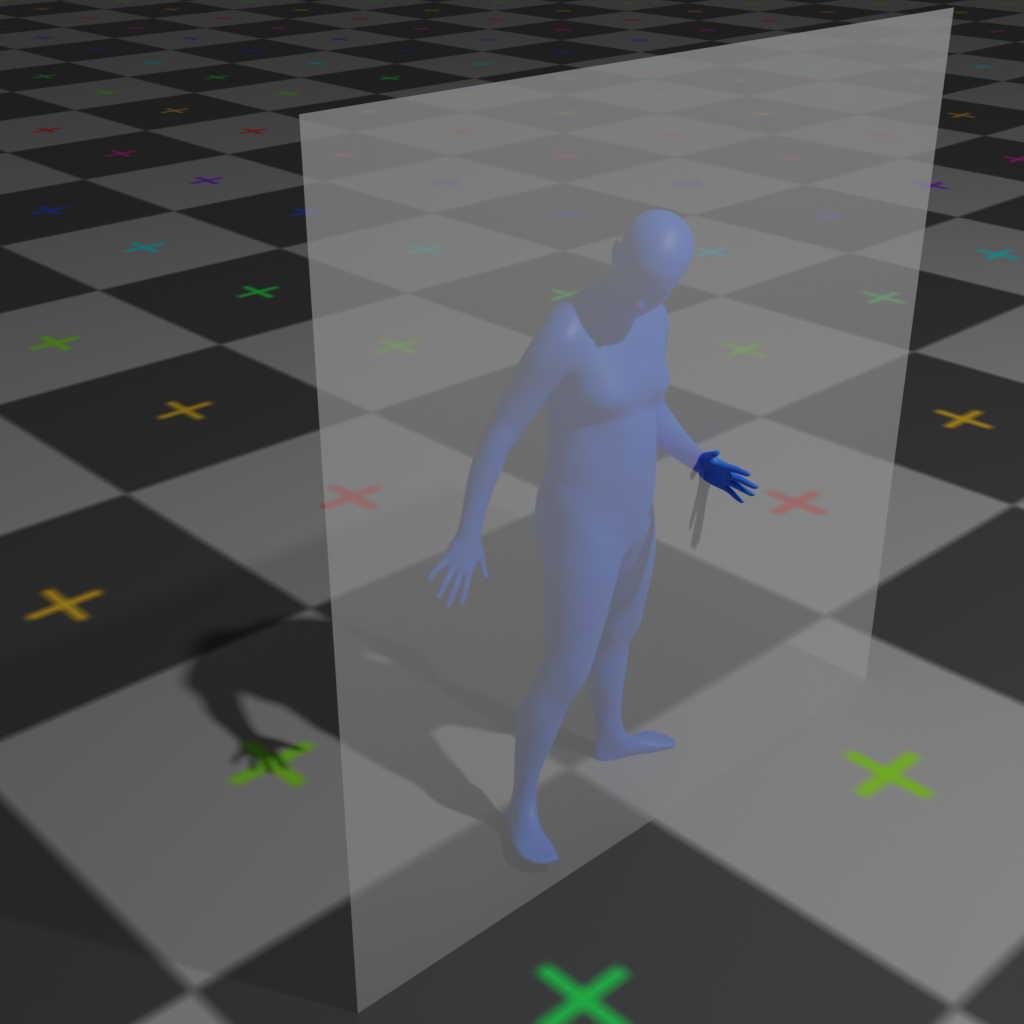} &
\includegraphics[trim={200px 0 100px 0}, clip, width=0.33\columnwidth]{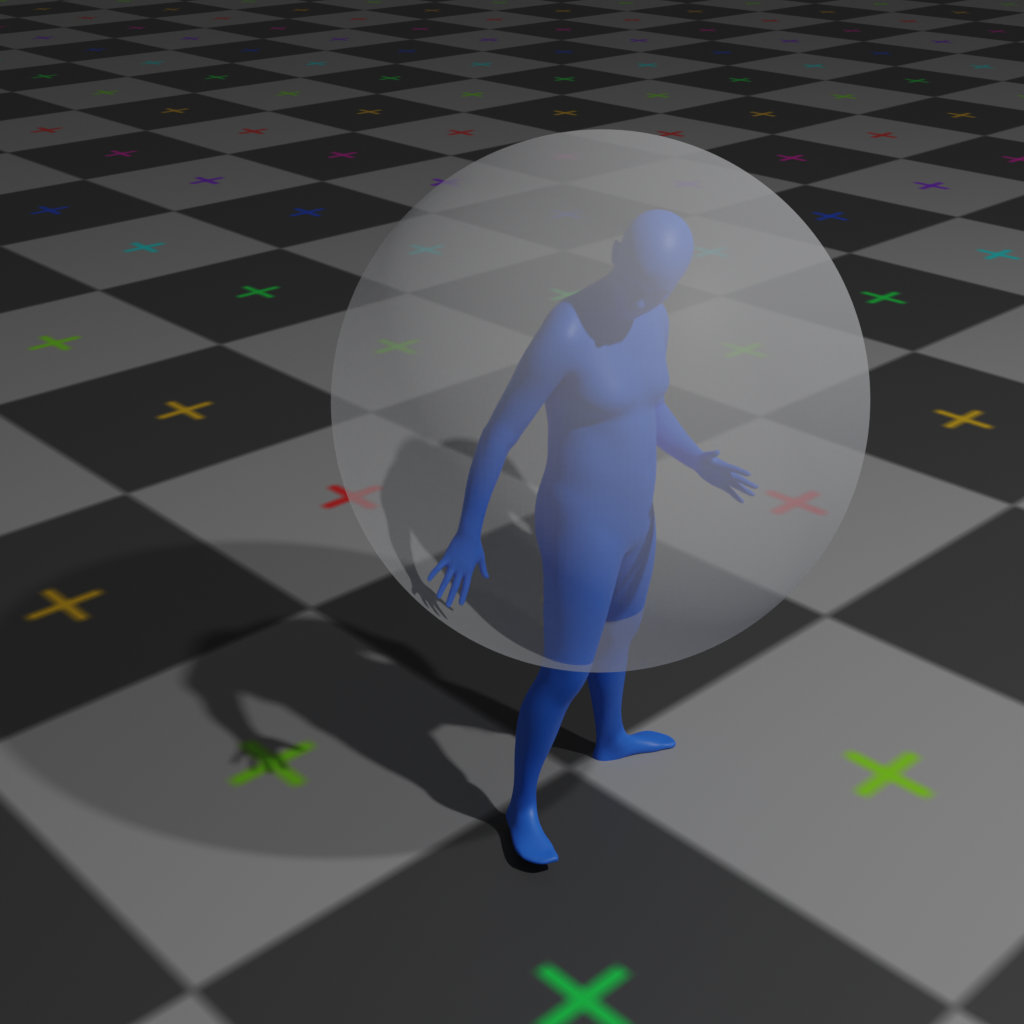} &
\includegraphics[trim={200px 0 100px 0}, clip, width=0.33\columnwidth]{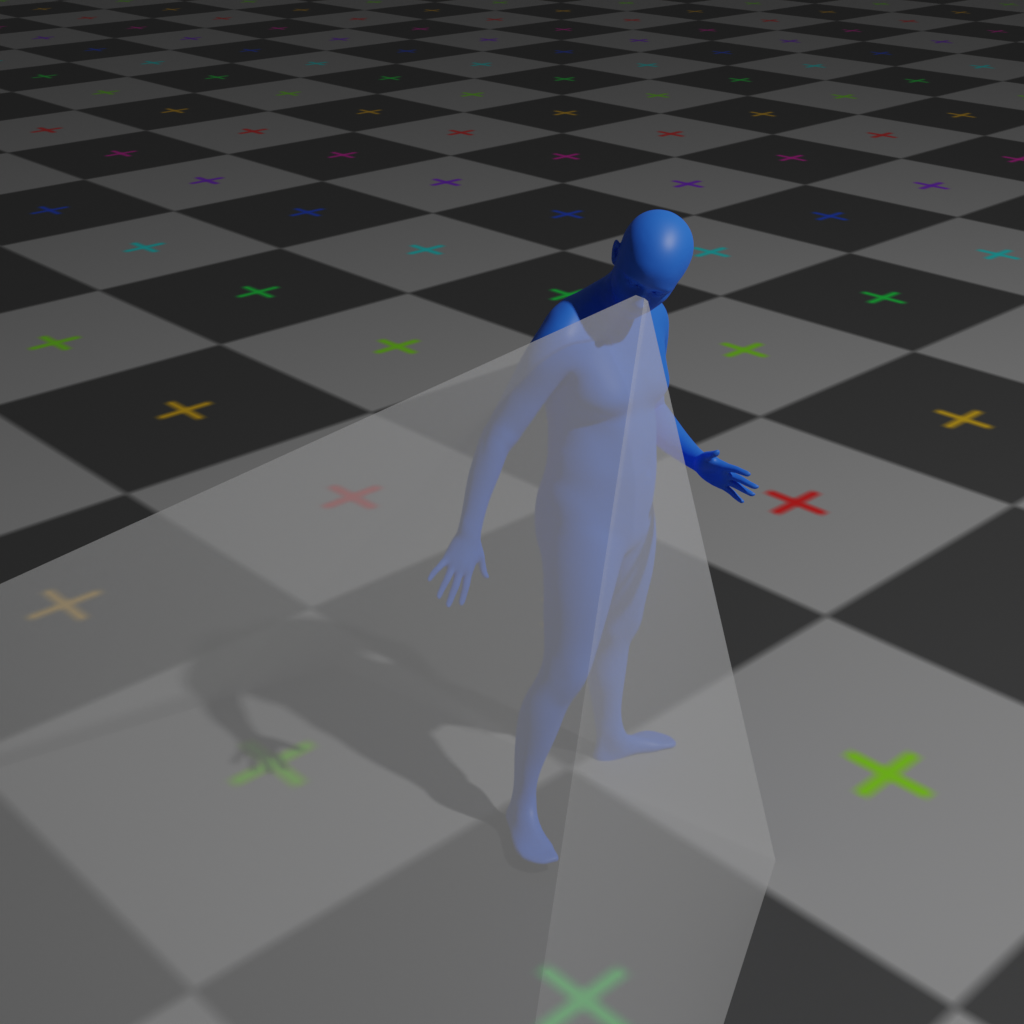} \\
(a) Plane FoV & (b) Motion Controllers (MC) & (c) Hand Tracking (HT)
\end{tabular}

\begin{tabular}{l c c }
\toprule
    Scenario & Left hand visibility & Right hand visibility \\
    \midrule
    Motion Controllers & 100\% & 100\% \\
    Hand Tracking & 53.45\%\  & 48.94\% \\
    \bottomrule
\end{tabular}

\caption{\textbf{Top}: Definition of FoV for different scenarios for the same pose. (a) Planar FoV, as used in~\cite{choutas2021learning}, wherein avatar's left hand is visible while the right hand is not. (b) Fully visible, as in motion controller scenarios, wherein both hands are always visible. (c) HMD's hand tracking camera's FoV, as in hand tracking scenarios, wherein avatar's right hand is visible while the left hand is not. \textbf{Bottom}: Hand visibility statistics on AMASS test set.}
\label{fig:fov}
\end{figure}

\section{Related Work}
\label{sec:related_work}
Over the past few years, different solutions have been proposed to the problem of full body pose generation given sparse or partial observations, such as  markers~\cite{zhang2021we,loper2014mosh,zanfir2021thundr,ghorbani2021soma}, images~\cite{kolotouros2021probabilistic,biggs20203d,zanfir2020weakly}, 2D joints/keypoints~\cite{pavlakos2019expressive,bogo2016keep}, Inertial Measurement Units (IMUs)~\cite{von2017sparse,huang2018deep,yi2021transpose,yi2022physical}, and HMDs~\cite{aliakbarian2022flag,ahuja2021coolmoves,yang2021lobstr,ponton2022mmvr,jiang2022avatarposer}. Among these works, the ones utilizing wearable devices are closer to our approach and thus we discuss them here.
Recently, different techniques have been proposed for human pose reconstruction using a sparse set of IMUs attached to the body~\cite{von2017sparse,huang2018deep,yi2021transpose,yi2022physical}. While attempts have been made to come up with a minimum number of IMUs for generating full body motion, typically an IMU sensor is used near pelvis, making the problem relatively easy compared to HMD scenarios where the root/pelvis signal is not available. 
Whilst external sensors, e.g., IMUs~\cite{von2017sparse,huang2018deep,yi2021transpose,yi2022physical} and cameras~\cite{saito2020pifuhd} are effective, they are often not as accessible as wearing a HMD. Using only a HMD is desirable from a usability point of view, but generating realistic and faithful human representations from such inputs remains technically challenging.

\begin{figure*}[!t]
\centering
\includegraphics[width=0.8\textwidth]{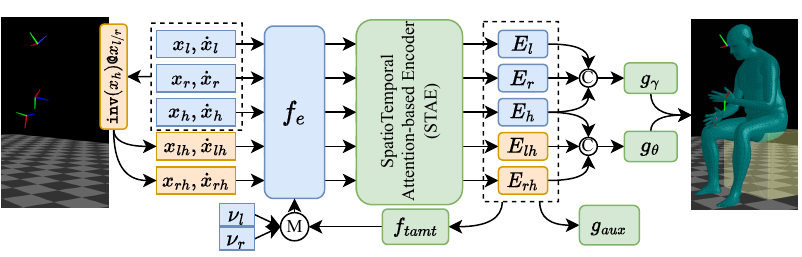}
\caption{Overview of HMD-NeMo: At each time-step, the model gets as input the HMD signal (left), as in Eq.~\ref{eq:input}. Such input is the mapped to an embedding space via $f_e$, acting as the input to the spatio-temporal encoder module STAE. The resulting feature representation is then utilized by two autoregressive decoders, $g\gamma$ and $g_\theta$ to predict the global trajectory as well as the pose, respectively. In case the model does not observe a hand (in HT scenario), $f_{tamt}$ fills in the representation of the unobserved hand effectively.}
\label{fig:method}
\end{figure*}

Given HMD signals,~\cite{aliakbarian2022flag} and~\cite{choutas2021learning} generate full body poses. While they generate expressive poses faithful to the HMD signal, these approaches~\cite{aliakbarian2022flag,choutas2021learning} only predict static poses, lacking the temporal consistency required for avatar motion generation. While~\cite{choutas2021learning} generates poses in the world space,~\cite{aliakbarian2022flag} predicts the pose relative to the root. Thus~\cite{aliakbarian2022flag} assumes a known root (pelvis joint) as an additional signal. The assumption of known root joint also appears in~\cite{dittadi2021full}, wherein, unlike~\cite{aliakbarian2022flag,choutas2021learning}, the method generates full body \emph{motion} given the HMD signal with Variational Autoencoders~\cite{kingma2013auto}. Note that, while~\cite{dittadi2021full} generates temporally plausible motions, it works in an offline fashion, predicting the motion for an entire sequence only after observing the whole sequence of HMD signals.
 
Recently, different techniques have been proposed to generate full body avatar locomotion (in the world coordinates) given HMD signals. In this context, \cite{ahuja2021coolmoves} proposes a matching algorithm, aiming to sample closest poses from a motion capture library at sparse time-steps and interpolate between poses. While this guarantees the selection of realistic poses, the output is always limited to the utilized motion capture library. In another work, \cite{yang2021lobstr} uses combination of an inverse kinematic (IK) solver and a recurrent neural network to generate upper body and lower body motions, respectively. Combination of different components to solve lower and upper body separately has also been explored in~\cite{ponton2022mmvr}, wherein a neural network is trained to predict the root orientation given the HMD signal, which then is used as the feature vector for a motion matching algorithm~\cite{buttner2015motion,holden2020learned} to generate full body. 
More recently, fully learning-based approaches have shown promise in generating full body avatar motion~\cite{winkler2022questsim,jiang2022avatarposer}. In this context, \cite{winkler2022questsim} simulates plausible and physically feasible motions within a reinforcement learning framework and \cite{jiang2022avatarposer} uses a transformer-based approach to generate full body motion given HMD signals.
 
Although great progress has been made by recent approaches, they all solve the problem of motion generation given head and \emph{both hands}, typically captured with motion controllers. However many mixed reality experiences do not have motion controllers available and instead rely on hand tracking from HMD mounted sensors (e.g., cameras). This introduces the challenge of hand tracking failures and partial hand visibility owing to the restricted field of views. To the best of our knowledge, despite its usability, motion generation in the presence of partial hand observation has not been well-explored. In this paper, we propose a method capable of generating high fidelity and plausible full body motion even in presence of partially visible hands.

\section{Proposed Method}
In this section, we first define the problem, the scenarios we consider, as well as the input and the desired output representation. We then present our proposed method.
\subsection{Problem Definition}
\label{sec:definition}

\noindent\textbf{Task.}
The task is to generate full-body 3D human locomotion (predicting both the instantaneous pose and the global trajectory of the human) given the sparse HMD signal in an \textit{online} fashion\footnote{Note that, similar to existing work on HMD-driven motion generation~\cite{ahuja2021coolmoves,jiang2022avatarposer,dittadi2021full}, our approach does not predict body shape parameters and only focuses on the poses.}. That is, given the input signal $x_t$ at each time-step $t$, the system should predict the 3D human pose and trajectory $y_t$ near-instantaneously. HMD-NeMo achieves this using a neural network parameterized by $\phi$.

\noindent\textbf{Scenarios.}
In this paper, we consider two scenarios: \textit{Motion Controllers (MC)} scenario, wherein hands are \emph{always} tracked via motion controllers using constellation tracking, as illustrated in Fig.~\ref{fig:fov} (b), and \textit{Hand Tracking (HT)} scenario, wherein hands are tracked via a visual hand tracking system whenever the hands are inside the FoV of the device, as illustrated in Fig.~\ref{fig:fov} (c). 
The FoV of the device is defined as a frustum determined by the HMD's hand tracking camera placement and parameters. In this paper, we use a similar FoV to that of the Microsoft HoloLens 2~\cite{ungureanu2020hololens}. Note that, while the MC scenario has been explored recently~\cite{ahuja2021coolmoves,yang2021lobstr,ponton2022mmvr,jiang2022avatarposer}, to the best of our knowledge, this work constitutes the first approach that tackles both MC and HT scenarios within one unified framework\footnote{Note that \cite{choutas2021learning} also considers a form of partial hand visibility by defining a plane FoV, wherein, as illustrated in Fig.~\ref{fig:fov} (a), joints in front of head is are considered visible. However, this scenario does not apply to practical use cases.}. The HT scenario is particularly challenging as hands tend to be out of FoV almost 50\% of the time, according to the statistics presented in the table in Fig.~\ref{fig:fov}.

\noindent\textbf{Input Representation.}
The input signal $x^t$ contains the head 6-DoF $x_h\in \mathbb{R}^{(6+3)}$, the left hand 6-DoF $x_{l}^t\in \mathbb{R}^{(6+3)}$, and the right hand 6-DoF $x_{r}^t\in \mathbb{R}^{(6+3)}$, all in the world space. We use the 6D representation to represent the rotations~\cite{zhou2019continuity}. We additionally provide the hand representations in the head space, $x_{lh}^t\in \mathbb{R}^{(6+3)}$ and $x_{rh}^t\in \mathbb{R}^{(6+3)}$. In the HT scenario, hands may go in and out of FoV of the HMD, so we also provide HMD-NeMo with the hand visibility status for both left and right hand, $\nu_l^t$ and $\nu_r^t$, as binary values, 1 being visible and 0 otherwise. Finally, for all 6-DoF signals, we provide the velocity of changes between two consecutive frames. Specifically, for translations we consider $vel(\mathcal{T}^t, \mathcal{T}^{t-1}) = \mathcal{T}^t - \mathcal{T}^{t-1}$, where $\mathcal{T}$ is the translation, and for rotations we consider the geodesic changes in the rotation $vel(R^t, R^{t-1})= (R^{t-1})^{-1}R^t$, where $R$ is the rotation, which together constitute the velocity 6-DoF $\dot{x}^t_.$. Overall, the input to the HMD-NeMo, $x^t\in\mathbb{R}^{92}$, can be written as
\begin{align}
x^t = \{ x_h, x_l, x_r, x_{lh}, x_{rh}, \dot{x}_h, \dot{x}_l, \dot{x}_r, \dot{x}_{lh}, \dot{x}_{rh}, \nu_l,\nu_r \}^t
\label{eq:input}
\end{align}

\noindent\textbf{Output Representation.}
The output of HMD-NeMo contains the pose (including the root orientation), $\theta^t\in \mathbb{R}^{J\times 3}$ represented with axis-angle rotations for the $J$ joints in the body, and the global position in the world, $\gamma^t\in \mathbb{R}^{3}$ represented as the root position, resulting in $y^t\in\mathbb{R}^{(J+1)\times 3}$,
\begin{align}
    y^t=\{\theta,\gamma \}^t
\end{align}
The sequence of $\theta^{0:T}$  and $\gamma^{0:T}$ then represent the avatar motion as well as its global trajectory for the period $[0, T]$. Note that, in this paper, we may drop the time superscript $t$ for better readability and include it when necessary.

\subsection{HMD-NeMo}
\label{sec:method}

\noindent\textbf{Overview.}
HMD-NeMo's pipeline is illustrated in Fig.~\ref{fig:method}. As described in Section~\ref{sec:definition}, the model gets as input the information about the head and hands in world coordinate system, the hands expressed in the head space, as well as their velocities, as described in Eq.~\ref{eq:input}. To express hands in head space, we consider $x_{lh} = x_h^{-1}x_{l}$ (similarly, $x_{rh} = x_h^{-1}x_{r}$). This representation then acts as the input to an embedding layer, $f_e$, which aims to (1) map the raw input to an embedding space and (2) handle the unobserved hands. Given the input $f_e$, the next module, STAE, learns (a) how each representation evolves over time and (b) how different components of the input, i.e, head and hands, are correlated. Once such a rich representation is obtained, two auto-regressive decoders, $g_\theta$ and $g_\gamma$, generate the body pose and the global position of the avatar, respectively. At each time step, the output of STAE is used to update the mask tokens (described below) as a representation for the hand signals that may be missing in the next time-step. To aid training, we also include an auxiliary task of human pose reconstruction in SE(3), denoted by $g_{aux}$.

\begin{figure}
\centering
\includegraphics[width=0.5\textwidth]{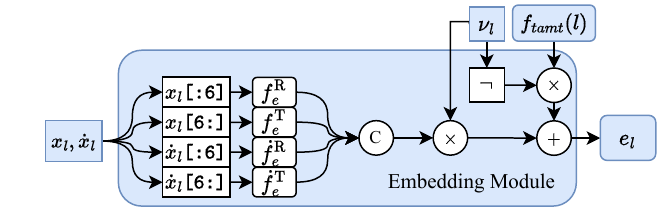}
\caption{Overview of the embedding module. Note that $x_l$ is only provided as an example and this module applies to all 6-DoF inputs appearing in $x^t$ as in Eq.~\ref{eq:input}.}
\label{fig:embedding}
\end{figure}

\noindent\textbf{Head and hand embedding ($f_e$).}
\label{subsec:embedding}
As shown in Fig.~\ref{fig:embedding}, the embedding module $f_e$ gets as input the head and hands 6-DoFs and velocities and maps them to a higher-dimensional embedding space. As the range of values corresponding to the rotations is different from those of the translations, we decouple such information and embed them via separate shallow MLPs and concatenate the results back together. For instance, for the observed left hand in the world coordinate system, the embedding representation is 
\begin{align}
e_l^{visible} = \bigg[f_e^\text{R}\big(x_l[:6]\big), f_e^\text{T}\big(x_l[6:]\big),\dot{f}_e^\text{R}\big(\dot{x}_l[:6]\big), \dot{f}_e^\text{T}\big(\dot{x}_l[6:]\big)\bigg]
\end{align}
where $f_e^\text{R}$ and $f_e^\text{T}$ are MLPs responsible for computing the rotation and translation embedding, acting on the first 6 elements (the 6D rotation representation) and the last 3 elements (the translation) of the input, respectively (similarly for $\dot{f}_e^\text{R}$ and $\dot{f}_e^\text{T}$ which act on velocities).

In the HT scenario, hands may not be visible to the model, hence computing such embedding representation is not possible. Thus, given the status of $\nu_l$ and $\nu_r$, the embedding module decides to either compute the embedding or utilize the output of the $f_{tamt}$ (described below), a set of temporally adaptable mask tokens, instead of a missing hand observation (denoted by $M$ in Fig.~\ref{fig:method}). As illustrated in Fig~\ref{fig:embedding}, the embedding of the left hand in the world coordinate system can be computed as
\begin{align}
e_l = \nu_l e_l^{visible} + (1-\nu_l)f_{tamt}(l).
\end{align}

\noindent\textbf{Spatio-temporal encoder (STAE).}
\label{subsec:encoder}
The output of $f_e$ on each component of the input stream is a non-temporal feature, computed independent of other components in the input. While an expressive representation of each component, it lacks temporal and spatial correlation information. We specifically care about these characteristics because the model is required to generate temporally coherent motion and also because the motion of one body part often impacts or determines the motion of other body part(s). To obtain a more informative representation from the head and hands, HMD-NeMo first learns the isolated temporal features of each component of input representation and then learns how they are spatially correlated~\cite{vaswani2017attention}. 

\begin{figure}
\centering
\includegraphics[width=0.5\textwidth]{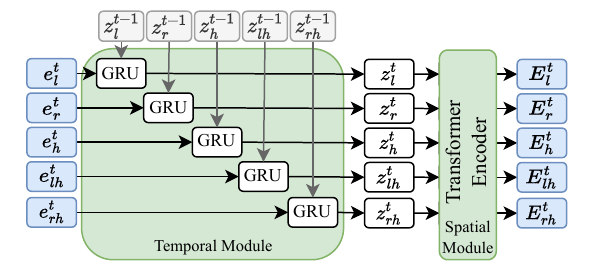}
\caption{Overview of the spatio-temporal encoder module. 
}
\label{fig:encoder}
\end{figure}

As illustrated in Fig.~\ref{fig:encoder}, to learn the temporal representation of the input signal, we use gated recurrent units (GRUs)~\cite{cho2014properties}. With a GRU module on top of each component in the input, the model learns how each component, e.g., head, evolves over time, independent of other components in the input. This information is compressed in the hidden state of the GRU cell, $z$, which is then utilized to learn how different components in the input relate to each other. This is achieved by using a standard transformer encoder on the GRU hidden states, thanks to the self-attention mechanism.

\begin{figure}
\centering
\includegraphics[width=0.5\textwidth]{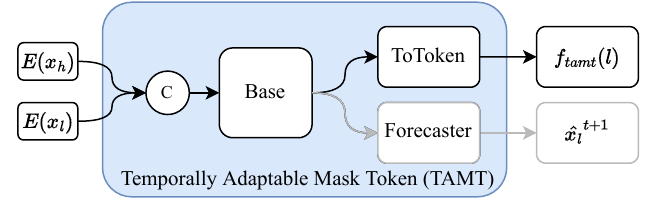}
\caption{Overview of TAMT module. 
Note that here we illustrate TAMT of $x_l$ as an example, the same applies f to both hands in both head space and world space.}
\label{fig:tamt}
\end{figure}

\noindent\textbf{Temporally adaptable mask tokens (TAMTs).}
\label{subsec:tamt}
As discussed in Section~\ref{sec:definition}, hands may not be visible to the model, and thus, there is no representative input signal for the $f_e$ module. To address this issue, in case of a missing hand observation, our model produces a feature vector, $f_{tamt}$, to represent the missing hand observation. To compute $f_{tamt}$, as illustrated in Fig.~\ref{fig:tamt}, we use the output of STAE for the hand observation that may be missing in the next time step as well as the output of STAE for the head. Note that head joint is the reference joint and is always available. The combination of these two features is a rich representation of the past state of the missing hand signal (both temporally and spatially); this is then used to compute the $f_{tamt}$. In order to encourage $f_{tamt}$ to learn information about the missing hand observation, as illustrated by the \emph{Forecaster} module in Fig.~\ref{fig:tamt}, we introduce a forecasting auxiliary task to forecast the state (6-DoF) of the corresponding hand in the next time-step.

\subsection{Training HMD-NeMo}
\label{subsec:training}

To train HMD-NeMo, we rely on the availability of a motion capture dataset, represented as SMPL~\cite{loper2015smpl} parameters (pose, shape, and global trajectory). For each sequence, we then simulate the HMD on the subject. In case of HT, we also simulate a FoV frustum to be able to model the hand visibility status ($\nu_l$ and $\nu_r$). We then train HMD-NeMo with a loss function of the form
\begin{align}
    \mathcal{L} = \alpha_{data}\mathcal{L}_{data} + \alpha_{smooth}\mathcal{L}_{smooth} 
    \\ \nonumber + \alpha_{SE(3)}\mathcal{L}_{SE(3)} + \alpha_{forecast}\mathcal{L}_{forecast} +  \alpha_{aux}\mathcal{L}_{aux} 
    \label{eq:total_loss}
\end{align}
The data loss term is the squared error between the predicted pose and trajectory and those of the ground truth,
\begin{align}
    \mathcal{L}_{data} = \sum_{t=1}^T||\hat{\theta}^t - \theta^t||_2^2 + ||\hat{\gamma}^t - \gamma^t||_2^2
\end{align}
Note that, in practice, the pose decoder has two heads, one each for predicting the body pose and the global root orientation. To further enhance the temporal smoothness, we penalize the discrepancy between the velocity of changes in the prediction to that of the ground truth
\begin{align}
    \mathcal{L}_{smooth} = \sum_{t=2}^T||\delta\hat{\theta}^t - \delta\theta^t||_1 + ||\delta\hat{\gamma}^t - \delta\gamma^t||_1
\end{align}
where $\delta\hat{\theta}^t=\hat{\theta}^{t} - \hat{\theta}^{t-1}$ ($\delta\hat{\gamma}$, $\delta\theta$, and $\delta\gamma$ follow similarly).
In addition to computing the reconstruction loss on the SMPL parameters, i.e., to relative joint rotations, we found it extremely useful to also utilize the reconstruction loss of each joint transformation independent of its parent, i.e., in the world space. To compute this reconstruction loss, we use the SMPL model to compute the joint transformations in SE(3) given the predicted and ground truth pose and trajectory parameters. Thus, the SE(3) reconstruction loss is 
\begin{align}
    \mathcal{L}_{SE(3)} = \sum_{t=1}^T||\hat{P}_{SE(3)}^t - P_{SE(3)}^t||_2^2
\end{align}
where $P_{SE(3)}$ is the body pose in SE(3). The next loss term corresponds to the forecasting auxiliary task in the TAMT module, where the goal is to minimize the distance between the predicted next hand and the ground truth next hand,
\begin{align}
    \mathcal{L}_{forecast} = \sum_{t=2}^T\sum_{j\in\{l,r\}}||\hat{x}^t_j - x^t_j||_2^2
\end{align}
Finally, we have our loss term for the auxiliary task, aiming to minimize the predicted full body joint transformations from STAE's features, $\hat{P}_{aux}$, to the ground truth body joint transformations,
\begin{align}
    \mathcal{L}_{aux} = \sum_{t=1}^T||\hat{P}_{aux}^t - P_{SE(3)}^t||_2^2
\end{align}
Our model is trained with $\alpha$s all being set to 1 in Eq.~\ref{eq:total_loss}.

\begin{figure*}
    \centering
    \scriptsize
\includegraphics[width=0.95\textwidth]{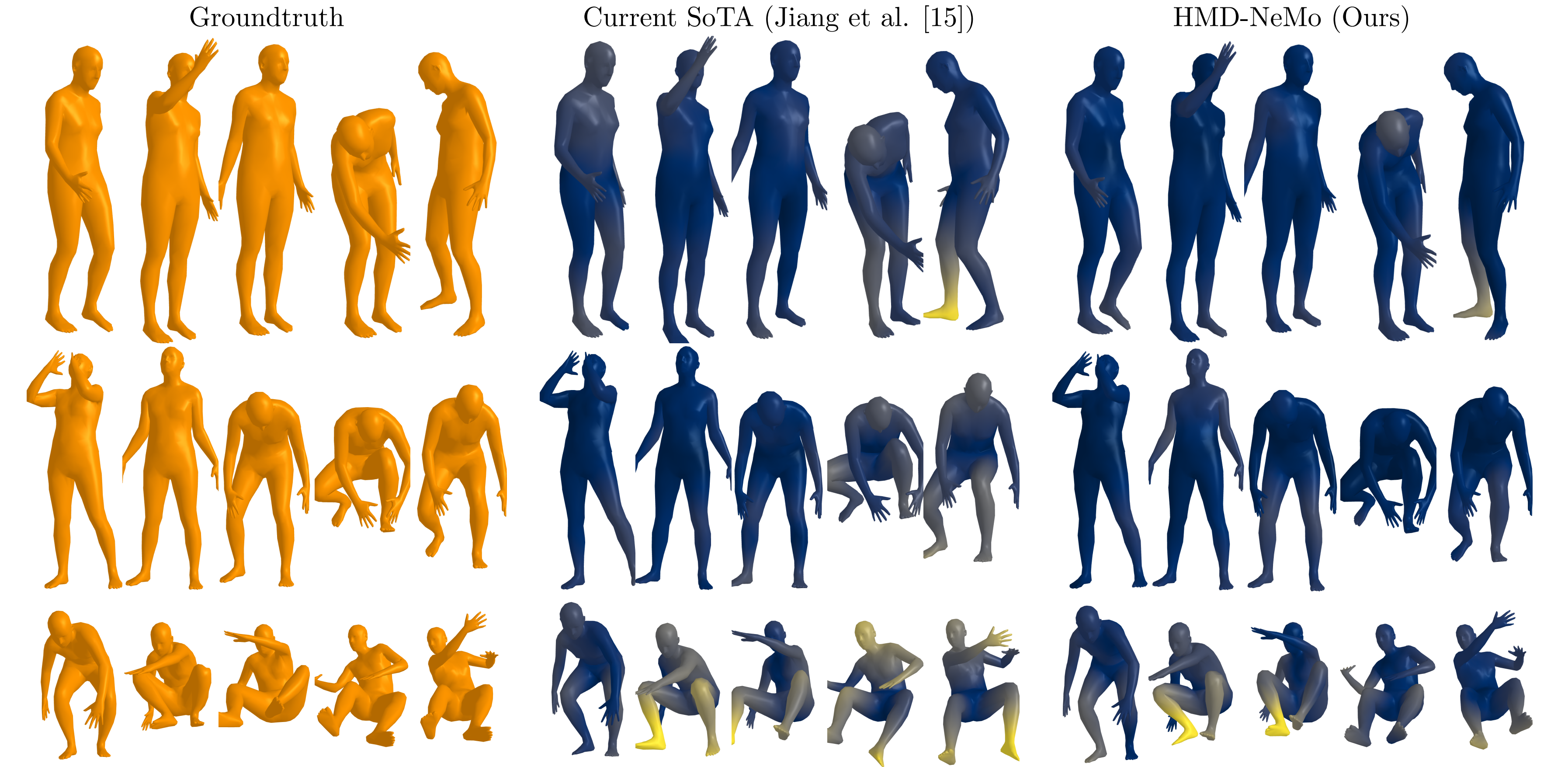}
\vspace{-5pt}
    \caption{Comparison to the state-of-the-art in MC scenario. Vertices are color-coded based on the distance to the GT (blue for low error and yellow for high error). Last row depicts a hard example with complex body pose and motion.}
    \label{fig:qualitative_MC}
\end{figure*}

\subsection{Optimization}
\label{sec:optimization}
Once trained, HMD-NeMo is capable of generating high fidelity and plausible human motion given only the HMD signal. However, as is typical of learning-based approaches, the direct prediction of the neural network does not precisely match the observations i.e., the head and hands, even if it is perceptually quite close. To close this gap between the prediction and the observation, optimization can be used. This adjusts the pose parameters to minimize an energy function of the form $\mathcal{E}=\mathcal{E}_{data}+\mathcal{E}_{reg}$, where $\mathcal{E}_{data}$ is the energy term that minimizes the distance between the predicted head and hands to the observed ones, and $\mathcal{E}_{reg}$ is additional regularization term(s). To define the data energy term, we define the residual $\mathcal{R} = \sum_{j\in \{h, l, r\}} (x_j - \hat{x_j})$, i.e., the difference between the predicted head/hand joint to that of the observation. Given $\mathcal{R}$, a typical, non-robust data energy term could be written as $\mathcal{E}_{nr} = \mathcal{R}^2$, i.e., the L2 loss. This suits the MC scenario perfectly, where head, left hand, and right hand are \textit{always} available. 
But this energy term may be misleading in HT scenario where hands are going into and out of FoV often, and thus leading to abrupt pose changes when hands appear back in the FoV (see supp mat for further details).
To remedy this issue, we utilize a more robust~\cite{barron2019general} alternative to the data energy term,
\begin{align}
\mathcal{E}_r(\mathcal{R}, a, b, c) = b \frac{|a-2|}{a} \Bigg( \bigg( \frac{(\frac{\mathcal{R}}{c})^2}{|a-2|} + 1 \bigg)^{(\frac{a}{2})} -1 \Bigg)
\label{eq:robust}
\end{align}
where $a$, $b$, and $c$ are hyper-parameters that determine the shape of the loss (see supp mat for further details). Unlike $\mathcal{E}_{nr}$, $\mathcal{E}_r$, considers large discrepancies between the prediction and observation as outliers, not penalizing the prediction strongly and thus does not push the prediction to move toward the observation. Thus abrupt changes in the arm poses are avoided and optimization stays on course despite large variation in the velocity metric (when caused by hand visibility changes). Of course, utilizing Eq.~\ref{eq:robust} adversely affects the fidelity, but a trade-off between the plausibility and fidelity can be chosen to suit the application of interest.

Note that, since all observations relate to the upper body, during optimization we only optimize the upper body pose parameters and global root trajectory, while keeping the predicted lower body untouched.

\section{Experiments}
\label{sec:experiments}
In this section, we introduce the dataset and metrics, with implementation details in the supplementary material. We then present the experimental results and ablation studies. 

\noindent\textbf{Dataset.}
We follow the recent common practice~\cite{jiang2022avatarposer,saleh2021probabilistic,choutas2021learning,dittadi2021full} of using a subset of AMASS~\cite{AMASS:ICCV:2019} for training and evaluation. AMASS is a large collection of human motion sequences, converted to 3D human meshes in the SMPL~\cite{loper2015smpl} representation, wherein every motion sequence contains information about poses $\theta$ and the global trajectory $\gamma$. To synthesize the HMD scenario, we compute the global transformation matrices for the head and hands as input. In the case of hand tracking, we define a FoV for the HMD and mask out the hands whenever they are out of FoV. To make a fair comparison, for both training and evaluation, we follow the splits suggested by~\cite{jiang2022avatarposer}.

\noindent\textbf{Metrics.}
To evaluate the performance of our approach as well as the competing baselines, we report the mean per-joint position error (MPJPE [cm]) and the mean per-joint velocity error (MPJVE [cm/s]). We compare our approach with recent techniques~\cite{jiang2022avatarposer,ahuja2021coolmoves,yang2021lobstr,dittadi2021full}\footnote{The approach~\cite{dittadi2021full} is modified to predict motion in world coordinates.}. As these approaches do not tackle the hand tracking scenario, we compare HMD-NeMo against them for the motion controller scenario. We evaluate HMD-NeMo for hand tracking scenarios separately. 
In our experiments, we do not use the ground truth body shape parameters, but instead use the same default shape for all sequences. This follows the evaluation used in previous work~\cite{ahuja2021coolmoves,jiang2022avatarposer,dittadi2021full}.

\begin{table}
\centering
\setlength{\tabcolsep}{10pt}
\setlength\extrarowheight{-3pt}
\begin{tabular}{l c c}
\toprule
Method & MPJPE $\downarrow$ & MPJVE $\downarrow$ \\
\midrule
FinalIK~\cite{finalIK}	& 18.09	& 59.24\\
Ahuja et al.~\cite{ahuja2021coolmoves}	& 7.83	& 100.54\\
Yang et al.~\cite{yang2021lobstr}	& 9.02	& 44.97\\
Dittadi et al.~\cite{dittadi2021full}	& 6.83	& 37.99\\
Jiang et al.~\cite{jiang2022avatarposer}	& 4.18	& 29.40\\
\midrule
HMD-NeMo (Ours)	& \textbf{1.90}	& \textbf{24.99}\\
\bottomrule
\end{tabular}
\caption{Comparison to the state-of-the-art approaches in MC scenario, where both hands are always visible.}
\label{tab:vr_comparison}
\end{table}

\begin{figure*}
    \centering
    \scriptsize
    \includegraphics[width=\textwidth]{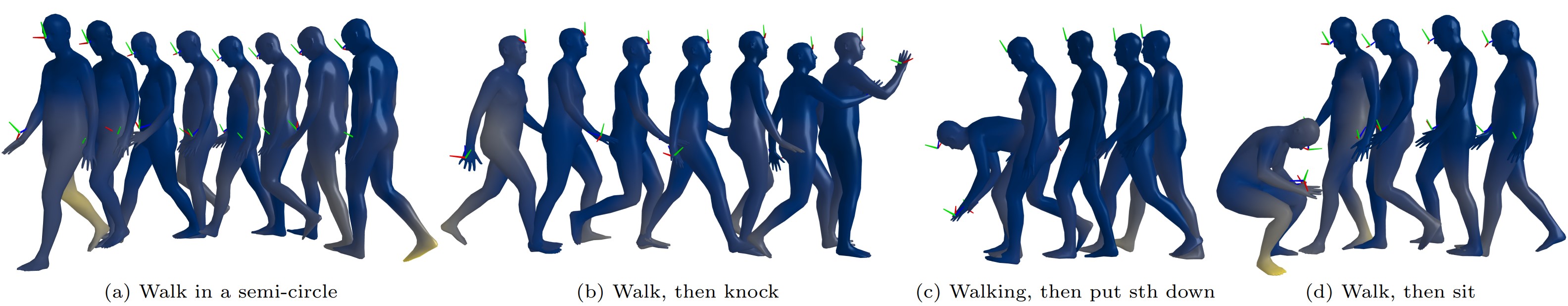}
    \caption{Qualitative results in HT scenario. Vertices are color-coded based on the distance to the GT (blue for low error and yellow for high error). See the supplementary video for more qualitative results.}
    \label{fig:qualitative_HT}
    \vspace{-5pt}
\end{figure*}

\subsection{Comparison to the state-of-the-art}
We compare HMD-NeMo with existing approaches that tackle the problem of full-body motion generation given HMD signal. To the best of our knowledge, all existing approaches only tackle the MC scenario, and thus we compare them with this setting\footnote{Note, we exclude very recent approaches~\cite{ponton2022mmvr,winkler2022questsim} since they either do not report on AMASS or the code is not publicly available.}.
As demonstrated in Table~\ref{tab:vr_comparison}, HMD-NeMo is not only more accurate (lower MPJPE), but also generates smoother motion with joint velocities similar to that in the ground truth (lower MPJVE), introducing a new state-of-the-art on the AMASS dataset for both metrics.
Note that \cite{finalIK} has the highest errors, as shown in Table~\ref{tab:vr_comparison} as it only optimizes the pose of the head and hands and ignores  the accuracy and smoothness of the rest of the joints. Consequently, since~\cite{yang2021lobstr} utilizes FinalIK within its framework, its performance becomes bounded by the quality of FinalIK. The learning-based approaches~\cite{jiang2022avatarposer,dittadi2021full}, however, perform much better than \cite{finalIK,ahuja2021coolmoves,yang2021lobstr}, highlighting the value of data-driven methods trained on large-scale motion capture datasets. In Fig.~\ref{fig:qualitative_MC}, we compared HMD-NeMo with the second best performing baseline~\cite{jiang2022avatarposer}.

\subsection{Ablation Studies}
\label{sec:ablation}

In this section, we comprehensively evaluate different aspects of HMD-NeMo, including evaluation in HT scenario, cross-dataset evaluations, ablation studies on input signals, model architectures, and loss terms. We also analyze the effect optimization in both HT and MC scenarios. Additionally, we provide results of HMD-NeMo in both HT and MC scenarios for various body parts, as well as qualitative effect of optimization in the supplementary material.

\noindent\textbf{Evaluation for the hand tracking scenario.}
To the best of our knowledge, HMD-NeMo is the first approach to address the problem of human motion generation given \emph{partial} HMD signal, applicable to HT scenario. We believe that TAMT module is the key to the success of HMD-NeMo in handling missing/partial observation, so we compare it with an alternative commonly used in the Vision Transformers~\cite{dosovitskiy2020image,he2022masked}, which uses a learned set of parameters (i.e., \texttt{nn.Parameters} in PyTorch) to model the missing observations\footnote{This can be considered as removing the unobserved hand (similar to~\cite{aliakbarian2022flag}) and replacing it with learned and fixed parameters.}. 
While learned parameters are fixed after training for every data point and every sequence, TAMT temporally updates itself at each time-step given the current state of the model. Table~\ref{tab:ar_comparison} (especially MPJVE) shows the superiority of TAMT module over learned and fixed parameters in handling missing hand observations.
Qualitatively, Fig.~\ref{fig:qualitative_HT} illustrates how HMD-NeMo performs in HT scenario.

\begin{table}
\small
\centering
\setlength\extrarowheight{-3pt}
\begin{tabular}{l l c c}
\toprule
Model & Mask Type & MPJPE $\downarrow$ & MPJVE $\downarrow$ \\
\midrule
HMD-NeMo & Learned \& fixed	& 5.59	& 42.81\\
HMD-NeMo & TAMTs (Ours)	&\textbf{ 2.48}	& \textbf{31.30}\\
\bottomrule
\end{tabular}
\caption{Effect of Temporally Adaptable Mask Tokens (TAMTs). These results represent the HMD-NeMo prediction before optimization solely to evaluate the effect of the TAMT module.}
\label{tab:ar_comparison}
\end{table}

\noindent\textbf{Cross-dataset evaluation.}
To investigate the generalizability of HMD-NeMo, we conduct a 3-fold cross-dataset evaluation as in~\cite{jiang2022avatarposer}, wherein the models are trained on two subsets and test on the other subset. In order to compare our approach with existing methods, we conduct this experiment in MC scenario. As shown in Table~\ref{tab:crossdataset}, HMD-NeMo outperforms existing approaches in all three datasets, by a considerable margin, highlighting its generalizability. 

\begin{table}[]

    \centering
    \scriptsize

\scalebox{0.9}{
\begin{tabular}{l @{} c@{ }c c@{ }c c@{ }c}
     \toprule
     Method & \multicolumn{2}{c}{Test on CMU} & \multicolumn{2}{c}{Test on BMLrub} & \multicolumn{2}{c}{Test on HDM05} \\
     \midrule
      & MPJPE $\downarrow$ & MPJVE $\downarrow$& MPJPE $\downarrow$ & MPJVE $\downarrow$& MPJPE $\downarrow$ & MPJVE $\downarrow$\\
      \midrule
FinalIK~\cite{finalIK}	        	&	18.82	&	56.83&	17.58	&	60.64&	18.43	&	62.39\\
Ahuja et al.~\cite{ahuja2021coolmoves}	      	&	18.77	&	139.17&	13.30	& 134.77&	17.90	&	140.61\\
Yang et al.~\cite{yang2021lobstr}      	        	&	12.96	&	49.94&	11.00	&	60.74&	11.94	&	48.26\\
Dittadi et al.~\cite{dittadi2021full}     	       	&	13.04	&	51.69&	9.69	&	51.80&	10.21	&	40.07\\
Jiang et al.~\cite{jiang2022avatarposer}    	&	8.37	&	35.76&	7.04	&	43.70&	8.05	&	30.85\\
\midrule
HMD-NeMo (Ours)                                & \textbf{7.13} &	\textbf{31.23}     & \textbf{6.46} &	\textbf{40.38 } & \textbf{6.80}	& \textbf{27.91}     \\
\bottomrule
\end{tabular}
}
    \caption{Results of cross-dataset evaluation between different methods. In order to compare with existing methods, we provide results in MC scenario.}
    \label{tab:crossdataset}
\end{table}

\noindent\textbf{Evaluating the effect of input signal.}
As discussed in Section~\ref{sec:definition}, HMD-NeMo utilizes head and hands, hands in the head space, as well as the corresponding velocities, as in Eq.~\ref{eq:input}. Table~\ref{tab:InputSignal} summarizes the effect of each component in the input signal. As shown, adding hands in the head space on top of head and hands in the world coordinates leads to better pose prediction, and thus reduces the MPJPE. Incorporating the velocities has a significant contribution to generating more temporally coherent motion, and thus reduces the MPJVE considerably. Considering all input signals, as in Eq.~\ref{eq:input} leads to best MPJPE and MPJVE.

\begin{table}[]
    \centering
    \small
    \setlength\extrarowheight{-3pt}
        \centering
        \begin{tabular}{l c c}
        \toprule
         Input Signal & MPJPE $\downarrow$ & MPJVE $\downarrow$ \\
         \midrule
         $x^t = \{x_h, x_l, x_r\}^t$ &	4.38 &	39.63 \\
         $x^t = \{x_h, x_l, x_r, x_{lh}, x_{rh}\}^t$ &	3.21 &	38.32\\
         $x^t = \{x_h, x_l, x_r, \dot{x}_h, \dot{x}_l, \dot{x}_r\}^t$ &	3.53 &	35.27\\
         \midrule
         Full input signals (Eq.~\ref{eq:input}) & 	\textbf{2.48} &	\textbf{31.30}\\
         \bottomrule
        \end{tabular}
        \caption{Evaluating the effect of various input signals in HT scenario. Note that the hand visibility status $\nu_l,\nu_r$ are always provided to the model.}
        \label{tab:InputSignal}
\end{table}

\noindent\textbf{Evaluating the effect of STAE.}
One core component of HMD-NeMo is the spatioTemporal attention-based encoder (STAE). Here, we study the design choices of STAE in Table~\ref{tab:STAE}. As shown, removing the GRU component, which is responsible to learn the temporal information of various input signals, affects the MPJVE considerably. Removing the transformer encoder, which aims at learning the relation between head and hands, adversely affects the MPJPE. Unsurprisingly, removing the entire STAE module, i.e., going from input embedding layer to the decoders, is a considerably weak baseline, with poor pose quality and temporal coherency. HMD-NeMo utilizes the power and efficiency of GRU module to learn the temporal information as well as the expressively of the transformer encoder to learn how various inputs are related to each other. This design leads to best performing MPJPE and MPJVE, especially in HT scenario wherein the model faces regular missing observations.

\begin{table}[]
    \centering
    \small
    \setlength\extrarowheight{-3pt}
        \centering
        \begin{tabular}{l c c}
        \toprule
         STAE Components & MPJPE $\downarrow$ & MPJVE $\downarrow$ \\
         \midrule
         No GRU & 4.34 &	46.61 \\
         No Transformer & 5.55 &	39.10\\
         No GRU, No Transformer & 7.09 &	52.26 \\
         \midrule
         With STAE (Ours) & \textbf{2.48} &	\textbf{31.30}\\
         \bottomrule
        \end{tabular}
        \caption{Evaluating the effect of STAE module in HT scenario.}
        \label{tab:STAE}
\end{table}

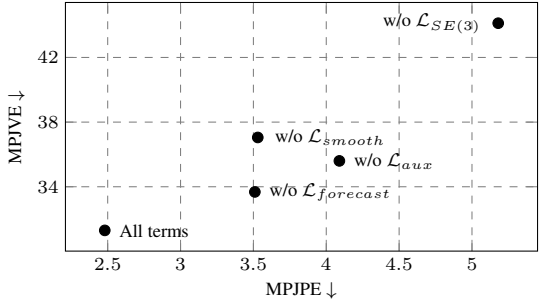
\begin{figure}
    \centering
    \pgfplotsset{width=0.45\textwidth, height=0.28\textwidth, compat=1.3, 
grid style={dashed,gray}}  
\begin{tikzpicture}
\scriptsize
\begin{axis}[
	ylabel={MPJVE $\downarrow$},
	xlabel={MPJPE $\downarrow$},
        ytick={30, 34, 38, 42, 46},
	grid=both,
]

\addplot [mark=*, mark size=2pt, black]
	coordinates {(2.48,	31.30)}
	node[right=3px] at (axis cs:2.48,	31.30) {\scriptsize All terms}; 

\addplot [mark=*, mark size=2pt, black]
	coordinates {(5.18,	44.12)}
	node[left=4px] at (axis cs:5.18,	44.12) {\scriptsize w/o $\mathcal{L}_{SE(3)}$}; 

\addplot [mark=*, mark size=2pt, black]
	coordinates {(3.53,	37.05)}
	node[right=4px] at (axis cs:3.53,	37.05) {\scriptsize w/o $\mathcal{L}_{smooth}$};
	
\addplot [mark=*, mark size=2pt, black]
	coordinates {(3.51,	33.68)}
	node[right=3px] at (axis cs:3.51,	33.68) {\scriptsize w/o $\mathcal{L}_{forecast}$};
	
\addplot [mark=*, mark size=2pt, black]
	coordinates {(4.09,	35.60)}
	node[right=3px] at (axis cs:4.09,	35.60) {\scriptsize w/o $\mathcal{L}_{aux}$}; 
	
\end{axis}
\end{tikzpicture}
    \caption{Evaluating the effect of each loss term. Please note that each term is evaluated in isolation.}
    \label{fig:loss_ablation}
\end{figure}

\noindent\textbf{Evaluating the effect of each loss term.}
As described in Section~\ref{subsec:training} and shown in Eq.~\ref{eq:total_loss}, HMD-NeMo is trained with five different loss terms. While $\mathcal{L}_{data}$ is the essential term for training HMD-NeMo, other loss terms contribute significantly to the performance of the model. In Fig.~\ref{fig:loss_ablation}, we illustrate the contribution of each loss term, in leave-one-term-out manner, on the changes in the MPJPE and MPJVE metrics. $\mathcal{L}_{smooth}$ has a significant impact on improving the MPJVE, while $\mathcal{L}_{aux}$ improves the MPJPE, and $\mathcal{L}_{forecast}$, which acts only on hands mildly improves MPJPE and MPJVE (its contribution to the total error metric is relatively small as it acts on a significantly fewer joints). $\mathcal{L}_{SE(3)}$ makes the largest contribution to the reduction of error in both metrics. It is likely that $\mathcal{L}_{SE(3)}$ aims to bridge that gap\footnote{This experiment only evaluates the contribution of each loss term independently. Evaluating the combination of loss terms remains for future investigations.} between the representation of the input signal (head and hand global transformation matrices) and the representation of the output pose (global root orientation and relative joint rotations). 

\noindent\textbf{Evaluating the effect of optimization.}
While the prediction of HMD-NeMo is very good, if the budget allows, it can be further optimized to improve accuracy (see supp mat for qualitative results). To bridge the gap between the predictions and observations, we optimize the pose prediction from HMD-NeMo, so that it matches the head and hand observations. As described in Section~\ref{sec:optimization}, in MC scenario, where both hands are always available, a simple optimization (with non-robust data energy term) loop can be used. The effect of such optimization in MC scenario is provided in Table~\ref{tab:vr_optimization}. However, considering the same strategy for the HT scenario, where we may partially observe hands, this may not be optimal as we lose plausibility in the generated motions (captured by the MPJVE metric). As described in Section~\ref{sec:optimization}, depending on the scenario and experience requirements, one may choose to (1) avoid optimizing the predictions if plausibility is the highest priority (first row of Table~\ref{tab:ar_optimization}), (2) use non-robust energy term if fidelity is the highest priority (second row of Table~\ref{tab:ar_optimization}), or (3) use a robust energy term as a trade-off between fidelity and plausibility (third row of Table~\ref{tab:ar_optimization}).

\begin{table}
\small
\centering
\setlength\extrarowheight{-3pt}
\begin{tabular}{l c c}
\toprule
Configuration & MPJPE $\downarrow$ & MPJVE $\downarrow$ \\
\midrule
Without optimization	& 2.07 	& 26.07 \\
With optimization	& \textbf{1.90}	& \textbf{24.99}\\
\bottomrule
\end{tabular}
\caption{Effect of optimization in MC scenario, where both hands are always visible.}
\label{tab:vr_optimization}
\end{table}

\begin{table}
\small
\centering
\setlength\extrarowheight{-3pt}
\begin{tabular}{l l c c}
\toprule
Configuration & Priority & MPJPE $\downarrow$ & MPJVE $\downarrow$ \\
\midrule
No Optimization	& Plausibility &2.48	& 31.30\\
Opt. with $\mathcal{E}_{nr}$ & Fidelity	& 2.32	& 33.51 \\
Opt. with $\mathcal{E}_{r}$	& Both & 2.37 & 31.33 \\
\bottomrule
\end{tabular}
\caption{Configuring HT scenario. Our model offers a range of choices to provide the best possible end-user experience despite incomplete observations.}
\label{tab:ar_optimization}
\end{table}

\noindent\textbf{Performance analysis.}
With 5.3M parameters, at inference time, HMD-NeMo requires only 4.4 ms to generate a pose given a HMD signal on a typical laptop CPU. On a NVIDIA Tesla P100 GPU, our model runs at 265 fps, with 1 iteration of optimization costs 3ms per frame (which could be further improved with more optimized implementation).
Such performance makes HMD-NeMo a potential solution for HMD-driven avatar animation in immersive environments.

\section{Conclusion}
In this paper, we present HMD-NeMo, a unified approach to generate full-body avatar motion in both motion controller and hand tracking scenarios. To handle the unobserved hands in a temporally coherent and plausible manner, we introduce TAMT module. It is worth noting that, in this paper, we considered one major reason for partial hand visibility, i.e., hands appearing out of the FoV, however, in practice, hands may not be visible to the hand tracking camera due to failure in tracking and occlusion by another object or by another body part. While such cases are not considered in the data augmentation in our paper, they certainly can be taken into account and TAMT can be used to fill such gaps with no additional modification; this is left for future exploration.
We provide extensive analyses on different components of HMD-NeMo, and shed light on various choices for optimization on top of neural network's predictions when it comes to production priorities (plausibility versus fidelity). 

\newpage
{\small
\bibliographystyle{ieee_fullname}
\bibliography{paper}
}

\pagebreak

\newpage
\section*{Supplementary Material}

\section*{Implementation Details}
To train HMD-NeMo, we utilize the Adam optimizer~\cite{kingma2014adam} with a batch size of 256 and a learning rate of $1e^{-3}$. 
We follow~\cite{jiang2022avatarposer} and train HMD-NeMo with sequences of length 40 frames, however, our approach can be used to generate sequences of arbitrary length at inference time. 
To optimize HMD-NeMo's prediction, we use limited-memory BFGS optimizer~\cite{liu1989limited}, with a history size of 10, learning rate of 1, and Strong-Wolfe line search function~\cite{nocedal2006numerical}.
We only optimize the upper body pose parameters as well as the global root trajectory as the observations (head and hands) represent upper body only. 
In the rest of this section, we describe the detailed design of each component of HMD-NeMo.

\paragraph{Head and hand embedding module.}
This module comprises four shallow MLPs per each 6-DoF of the input: a MLP to compute the rotation representation, a MLP to compute the translation representation, a MLP to compute the rotational velocity representation, and finally a MLP to compute the positional velocity representation. Each MLP is a single \texttt{Linear} layer followed by \texttt{LeakyReLU} non-linearity. Each MLP maps its input (either rotation in 6D or translation in 3D) to a vector of size 32 in the latent space. For each 6-DoF representation (i.e., head, left hand, right hand, left hand in the head space, and right hand in the head space),  the result of the four MLPs are then concatenated together to form a vector of size 128.

\paragraph{SpatioTemporal encoder (STAE).}
This module comprises two sub-modules: a GRU-based module to encode temporal information and a transformer-based module to encode spatial information. Given the embedding representation of each input 6-DoF (of size 128), we consider a single-layer \texttt{GRU} with the hidden size of 256 to process each input signal temporally. The hidden state of each GRU cell is updated given its input and the previous hidden state. For each GRU, we initialize the hidden state at time $t=0$ with a MLP (a \texttt{Linear} layer followed by \texttt{Tanh} non-linearity) that gets as input the head embedding at time $t=0$ (head is considered the reference joint and it is always visible) and computes the initial hidden state. For each input representation, we have a separate GRU layer and a separate hidden state initialization MLP.
Since we have five 6-DoFs in the input signal, thus we compute and update five separate hidden states of the GRU. Such hidden states encompass the temporal information about each component of the input separately. At each time-step, these hidden states are then used as the input to a transformer encoder to learn how these temporal features are spatially correlated to each other. Specifically, we use 4 layers of transformer encoder, each with 4 attention heads and a feed-forward hidden dimension of 512.

\paragraph{Temporally adaptable mask tokens (TAMT).}
In order to take care of missing observations for hands, where computing the hand embedding representations is not feasible, we introduce TAMT, as described in the main paper.
For each hand, TAMT contains a base MLP (two layers of \texttt{Linear-LeakyReLU}) which gets as input the concatenation of the head representation and the corresponding hand representation, computed by the transformer encoder at time $t$. The output of the base MLP is then passed as input to two separate MLPs: a MLP (a single \texttt{Linear} layer followed by \texttt{LeakyReLU} non-linearity), called ToToken, that computes a vector of size 128 (same size as the output of head and hand embedding module) that produces TAMT features for time $t+1$, and a MLP (a \texttt{Linear} layer, followed by \texttt{LeakyReLU} non-linearity, followed by another \texttt{Linear} layer), called Forecaster, that computes/forecasts the 6-DoF of the corresponding hand in time $t+1$. The base MLP produces a feature vector of size 256 and Forecaster module's intermediate hidden dimension is also 256.
For the very initial time-step, if a hand observation is missing, we use a learned parameters for TAMT (learned via \texttt{nn.Parameters}).

\section*{Additional Ablation Studies}

\subsection*{Robust energy term}

\paragraph{Why we need a robust energy term?}
As described in the main paper, once trained, HMD-NeMo is capable of generating high fidelity and plausible human motion given only the HMD signal. However, as is typical of learning-based approaches, the direct prediction of the neural network does not precisely match the observations i.e., the head and hands, even if it is perceptually quite close. To close this gap between the prediction and the observation, optimization can be used. This adjusts the pose parameters to minimize an energy function of the form $\mathcal{E}=\mathcal{E}_{data}+\mathcal{E}_{reg}$, where $\mathcal{E}_{data}$ is the energy term that minimizes the distance between the predicted head and hands to the observed ones, and $\mathcal{E}_{reg}$ is additional regularization term(s). To define the data energy term, we define the residual $\mathcal{R} = \sum_{j\in \{h, l, r\}} (x_j - \hat{x_j})$, i.e., the difference between the predicted head/hand joint to that of the observation. Given $\mathcal{R}$, a typical, non-robust data energy term could be written as $\mathcal{E}_{nr} = \mathcal{R}^2$, i.e., the L2 loss. This suits the MC scenario perfectly, where head, left hand, and right hand are \textit{always} available. 
But this energy term may be misleading in HT scenario where hands are going into and out of FoV often, and thus leading to abrupt pose changes when hands appear back in the FoV. Specifically, consider that the right hand was out of FoV for a relatively long period of time up to time $t$ and the model has predicted what the right arm motion could be like for this period. Then, at time $t$, the right hand comes back to the FoV and thus we have an observed right hand signal. 
While the motion generated by the model is plausible, the predicted right hand may end up in a completely different location from the newly observed right hand. If we use the data energy term $\mathcal{E}_{nr}$ to minimize the total energy during optimization, we end up in an abrupt jump in the right arm pose from time $t-1$ to time $t$. 
While this guarantees high fidelity, i.e., hands in the correct position once observed, it adversely affects the perceptual experience of generating temporally smooth and coherent motion. 

\begin{figure}[!h]
    \centering
         \includegraphics[width=0.31\textwidth]{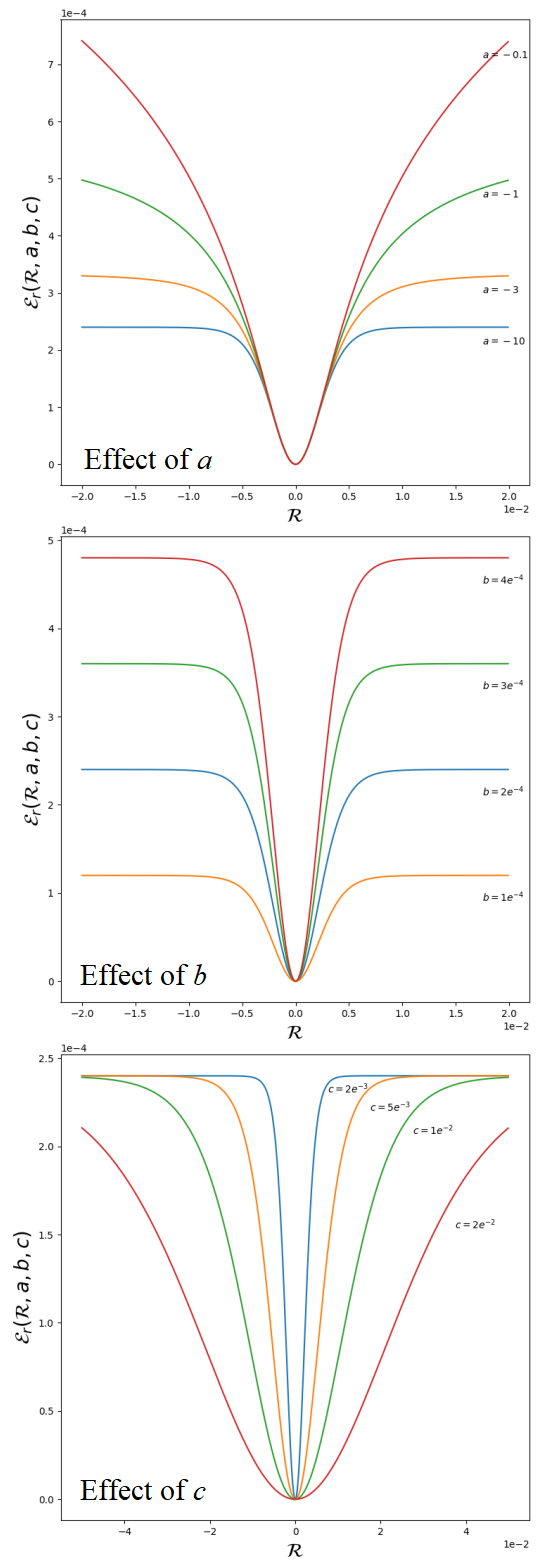}
    \caption{Effect of each hyper-parameter in determining the shape of the robust energy term, Eq.~\ref{eq:robust}.}
    \label{fig:robust_hyper}
\end{figure}

\begin{figure*}
    \centering
    \includegraphics[width=\textwidth]{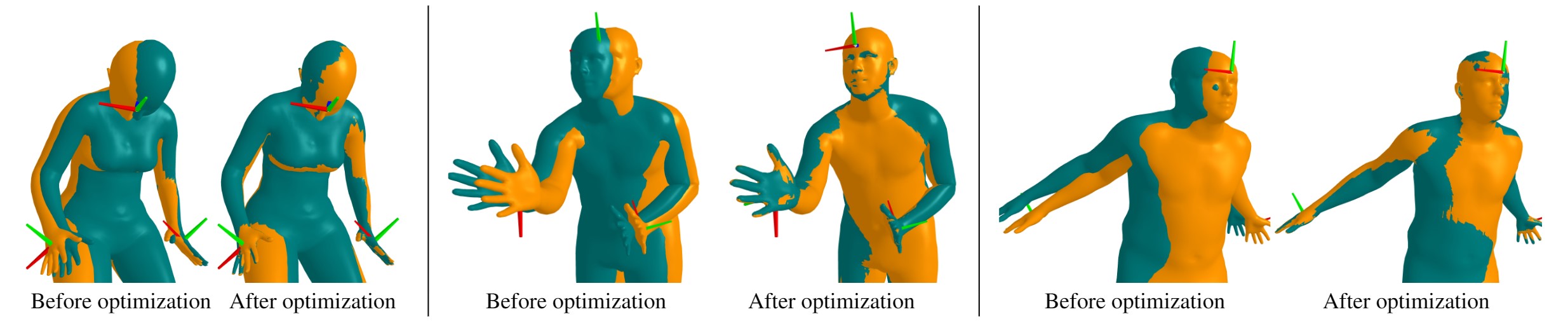}
    \caption{Illustration of the effect of optimizing HMD-NeMo predictions (predictions are shown in teal and GT in orange, overlaid). The initial model prediction is relatively accurate, but just a single optimization iteration improves the head and hands prediction substantially.}
    \label{fig:opt_vis}
\end{figure*}

\paragraph{Parameters of robust energy term.}
Inspired by the general form a robust loss function~\cite{barron2019general}, in this paper we use such technique to define our robust energy term for the optimization in the hand tracking scenario. In hand tracking scenario, where hands may appear outside of the field of view of the HMD, a non-robust energy term, e.g., L2, is ideal for the fidelity (hand poses appearing accurately when there is a hand observation), while may not be ideal for plausibility, as a result of abrupt jumps when a new hand signal is observed. This is why a robust alternative
\begin{align}
\mathcal{E}_r(\mathcal{R}, a, b, c) = b \frac{|a-2|}{a} \Bigg( \bigg( \frac{(\frac{\mathcal{R}}{c})^2}{|a-2|} + 1 \bigg)^{(\frac{a}{2})} -1 \Bigg)
\label{eq:robust}
\end{align}
is used when plausibility in the generated motions is a priority. As described in the main paper, the values of the hyper-parameters $a$, $b$, and $c$ affect the shape and thus the behaviour of the energy term. Particularly, such hyper-parameters determine (1) what range of values of $\mathcal{R}$ should be considered outlier, and thus not being penalized strongly, and (2) what is the penalty strength for the inliers and outliers. The effect of each parameter is visualized in Fig.~\ref{fig:robust_hyper}. Particularly, parameter $a$ determines the strength of the penalty as outliers go further from inlier region. Parameter $b$ determines the value of the loss at which it considers outlier. Parameter $c$ determines width at which we consider $\mathcal{R}$ as inlier. In our experiments, $a=-10$, $b=2e^{-4}$, and $c=2e^{-3}$.
Numbers on the plots of Fig.~\ref{fig:robust_hyper} are best seen when zoomed in. Fig.~\ref{fig:opt_vis} illustrates multiple examples before and after only 1 iteration of optimization.

\subsection*{Evaluation on various body parts}
In Table~\ref{tab:part_ablation} we compare the MC and HT scenarios, and break down the errors for various body parts. Since the HMD signal represents the upper body, the contribution of lower body joints towards the error (both MPJPE and MPJVE) is larger than that of the upper body joints. As expected, head and hand errors are relatively low since HMD signals represent head and hands. Also, as expected, the results in Table~\ref{tab:part_ablation} demonstrate that the motion prediction task in HT scenario is more difficult than in MC. Despite being user-friendly, HT scenario has not been well-explored by the community yet due to the technical difficulty of motion prediction in this setting. 

\begin{table}[]
\small
    \centering
    \setlength\extrarowheight{-1.5pt}
    \begin{tabular}{l l c c}
    \toprule
     &  & Motion  & Hand  \\
    Setting & Metric &  Controllers &  Tracking \\
    \midrule
    \multirow{2}{*}{Full body}
    & { MPJPE $\downarrow$} & 2.07 & 2.48\\ 
    & { MPJVE $\downarrow$} & 26.07 & 31.30\\
    \midrule
    \multirow{2}{*}{Upper body} 
    & { UB-MPJPE $\downarrow$} & 1.87 & 2.28\\
    & { UB-MPJVE $\downarrow$} & 24.26 & 29.76\\
    \midrule
    \multirow{2}{*}{Lower body} 
    & { LB-MPJPE $\downarrow$} & 2.56 & 2.80\\
    & { LB-MPJVE $\downarrow$} & 30.77 & 33.03\\
    \midrule
    \multirow{2}{*}{Head \& Hands} 
    & { HH-MPJPE $\downarrow$} & 0.88 & 1.72\\
    & { HH-MPJVE $\downarrow$} & 13.83 & 28.74\\
    \bottomrule
    \end{tabular}
    \caption{Per body part evaluation. Note these results represent the HMD-NeMo prediction prior to optimization.}
    \label{tab:part_ablation}
\end{table}

\section*{Additional Qualitative Results}
\begin{figure*}
    \centering
    \includegraphics[width=\textwidth]{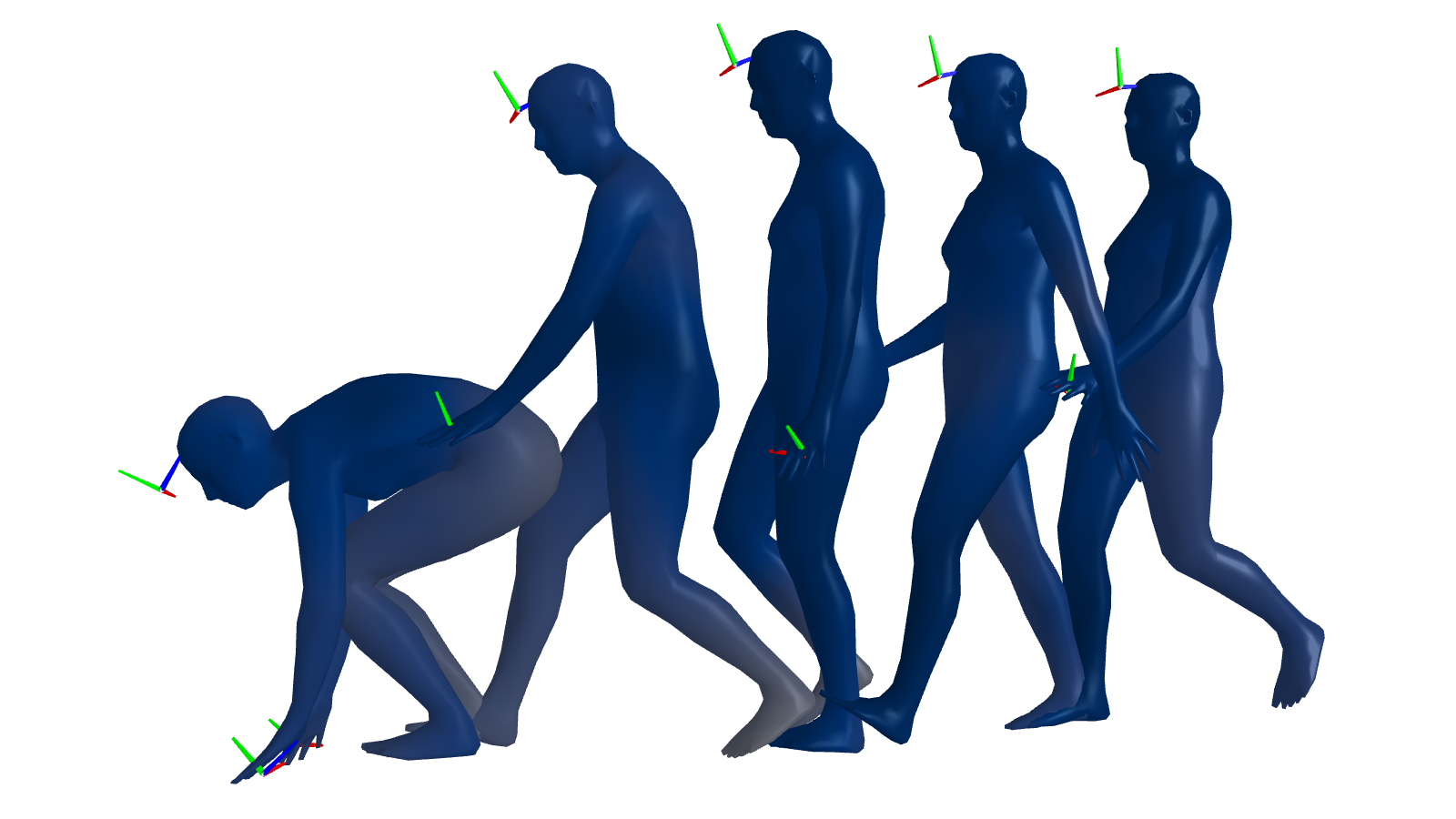}
    \caption{Qualitative results of HMD-NeMo in HT scenario.}
    \label{fig:supp_vis1}
\end{figure*}

\begin{figure*}
    \centering
    \includegraphics[width=\textwidth]{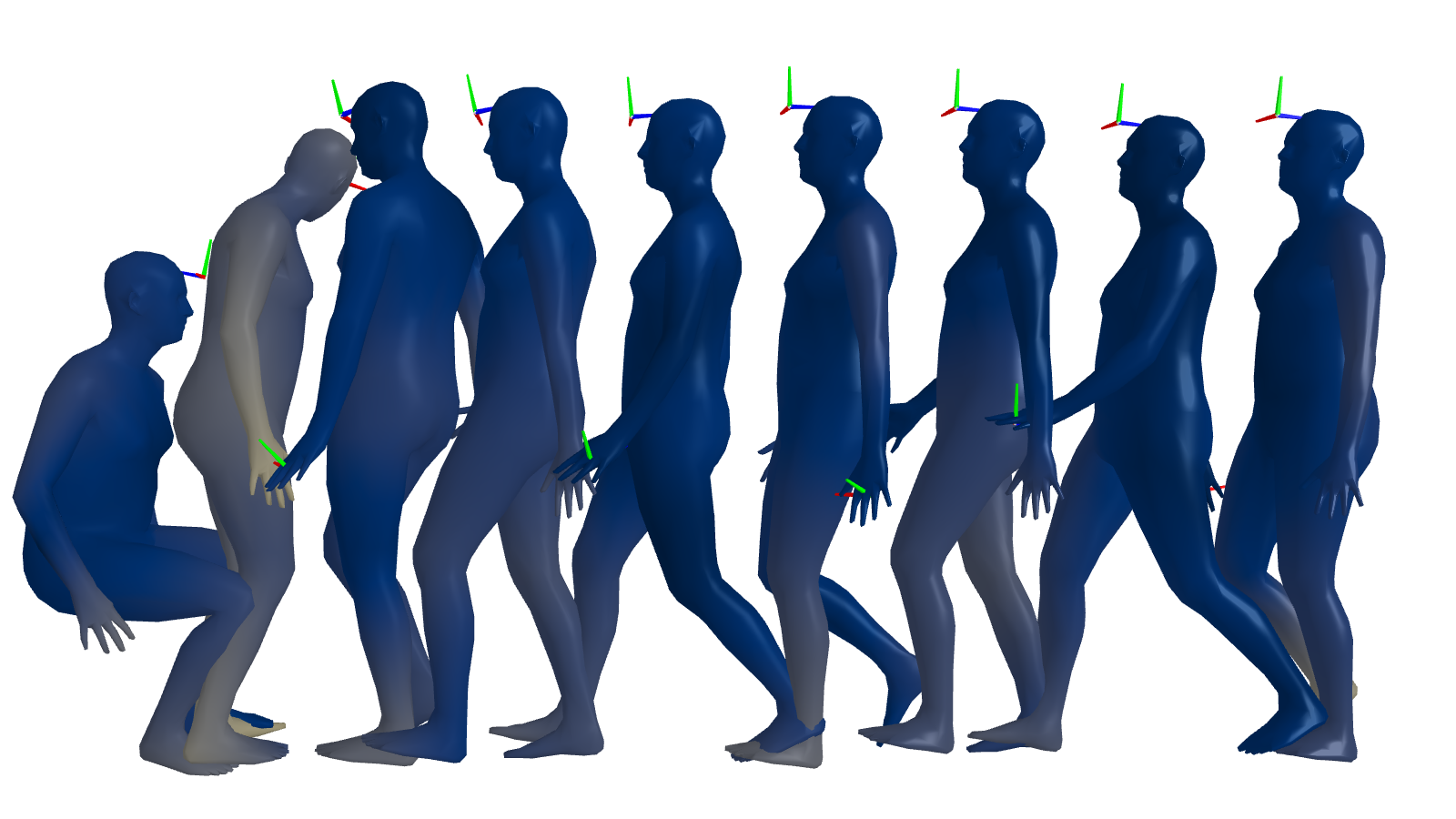}
    \caption{Qualitative results of HMD-NeMo in HT scenario.}
    \label{fig:supp_vis2}
\end{figure*}

\begin{figure*}
    \centering
    \includegraphics[width=\textwidth]{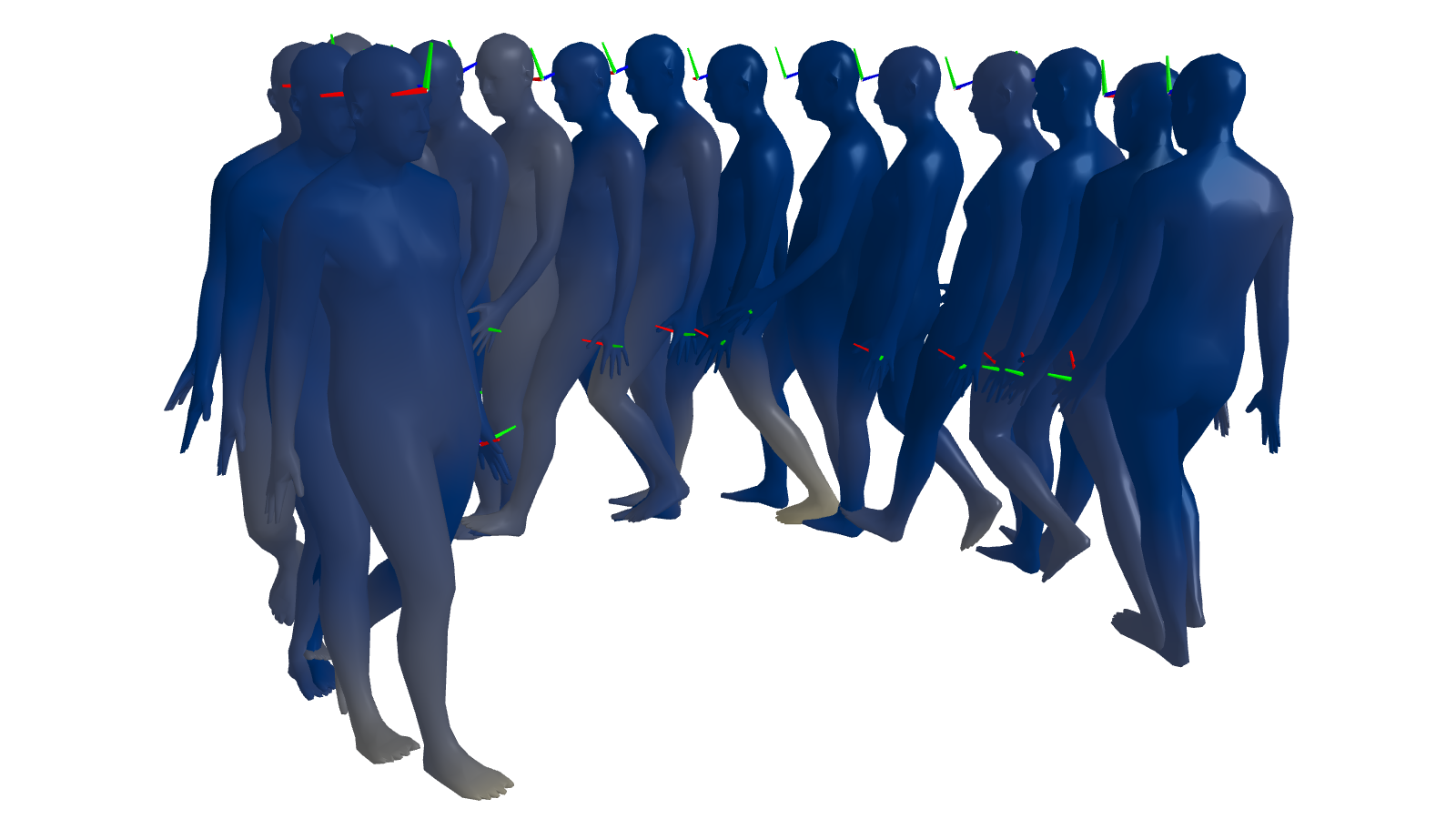}
    \caption{Qualitative results of HMD-NeMo in HT scenario.}
    \label{fig:supp_vis3}
\end{figure*}

\begin{figure*}
    \centering
    \includegraphics[width=\textwidth]{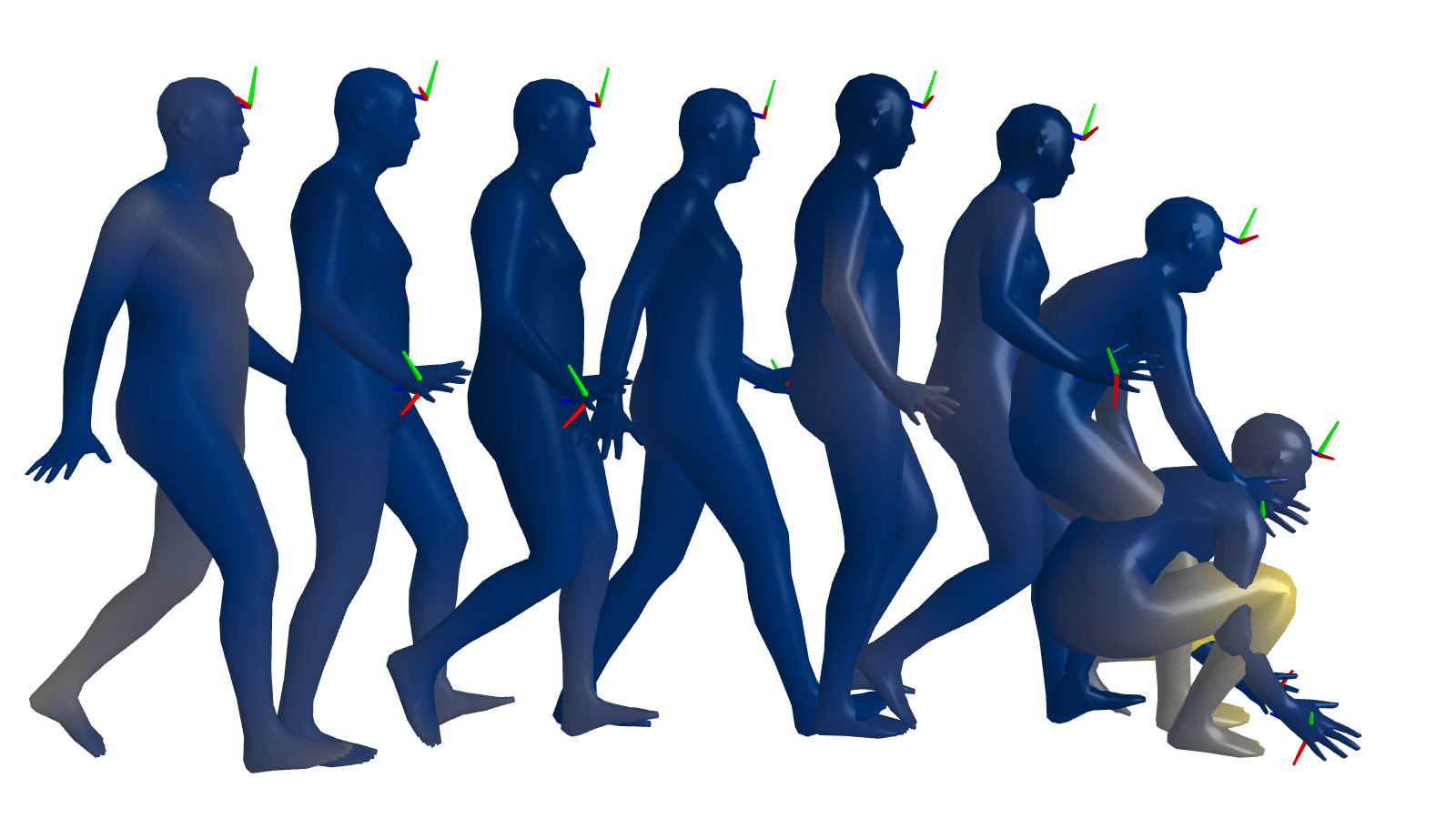}
    \caption{Qualitative results of HMD-NeMo in HT scenario.}
    \label{fig:supp_vis4}
\end{figure*}

\begin{figure*}
    \centering
    \includegraphics[width=\textwidth]{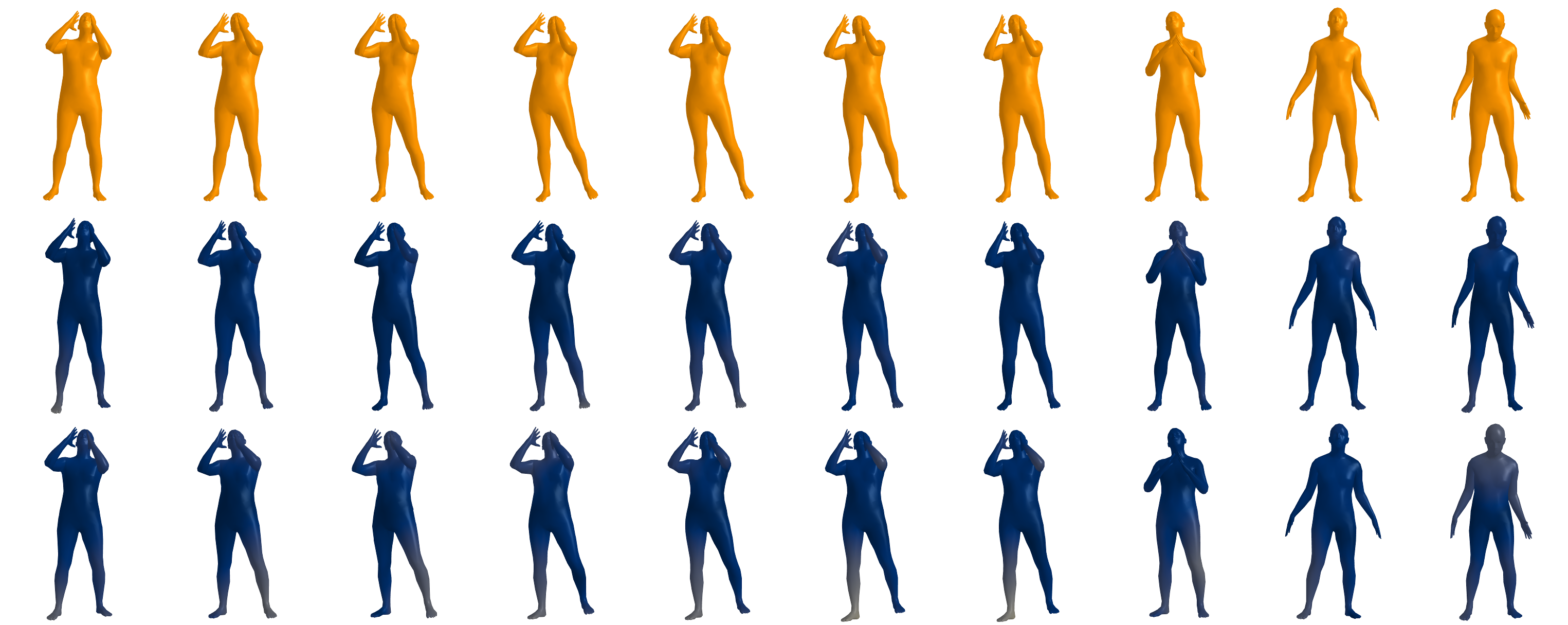}
    \caption{Qualitative comparison to the state of the part method~\cite{jiang2022avatarposer} in MC scenario. \textbf{Top}: Ground truth in orange, \textbf{Middle}: HMD-NeMo, \textbf{Bottom}: Jiang et al.~\cite{jiang2022avatarposer}.}
    \label{fig:supp_vis5}
\end{figure*}

\begin{figure*}
    \centering
    \includegraphics[width=\textwidth]{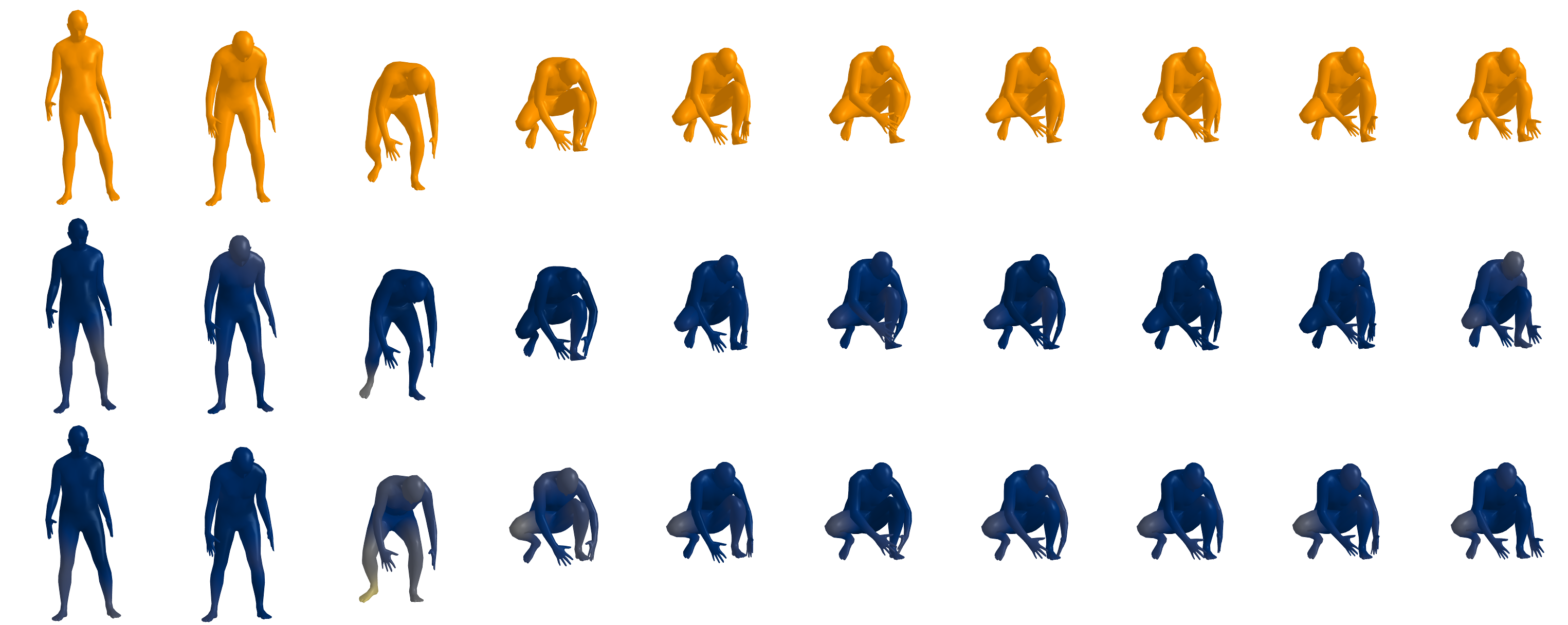}
    \caption{Qualitative comparison to the state of the part method~\cite{jiang2022avatarposer} in MC scenario. \textbf{Top}: Ground truth in orange, \textbf{Middle}: HMD-NeMo, \textbf{Bottom}: Jiang et al.~\cite{jiang2022avatarposer}.}
    \label{fig:supp_vis6}
\end{figure*}

\begin{figure*}
    \centering
    \includegraphics[width=\textwidth]{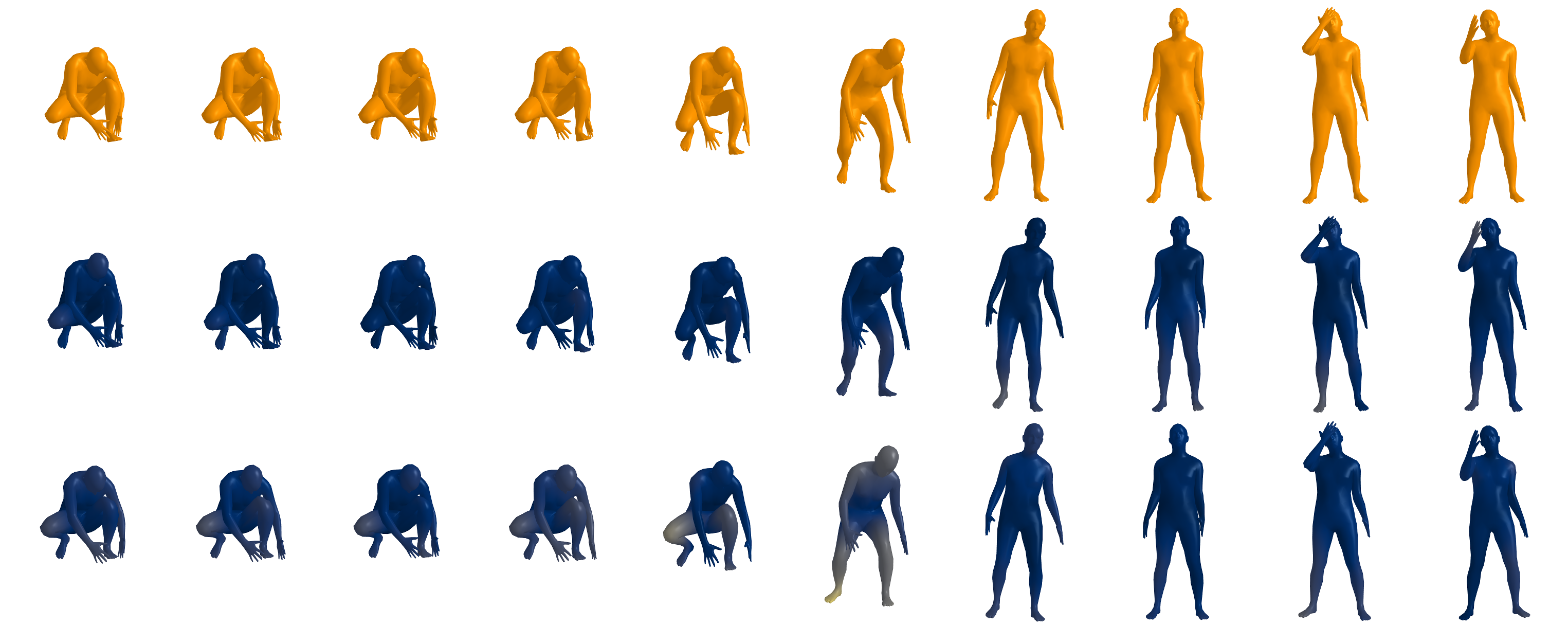}
    \caption{Qualitative comparison to the state of the part method~\cite{jiang2022avatarposer} in MC scenario. \textbf{Top}: Ground truth in orange, \textbf{Middle}: HMD-NeMo, \textbf{Bottom}: Jiang et al.~\cite{jiang2022avatarposer}.}
    \label{fig:supp_vis7}
\end{figure*}

\begin{figure*}
    \centering
    \includegraphics[width=\textwidth]{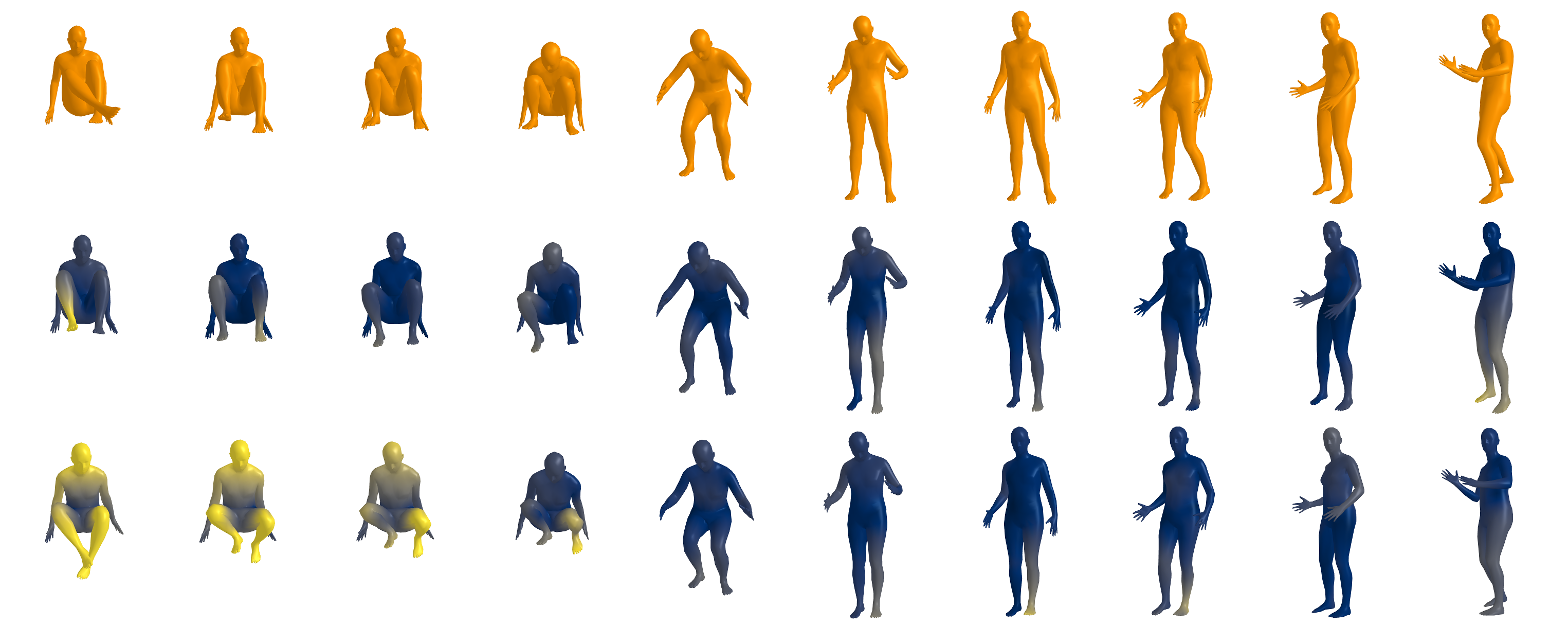}
    \caption{Qualitative comparison to the state of the part method~\cite{jiang2022avatarposer} in MC scenario. \textbf{Top}: Ground truth in orange, \textbf{Middle}: HMD-NeMo, \textbf{Bottom}: Jiang et al.~\cite{jiang2022avatarposer}.}
    \label{fig:supp_vis8}
\end{figure*}

\begin{figure*}
    \centering
    \includegraphics[width=\textwidth]{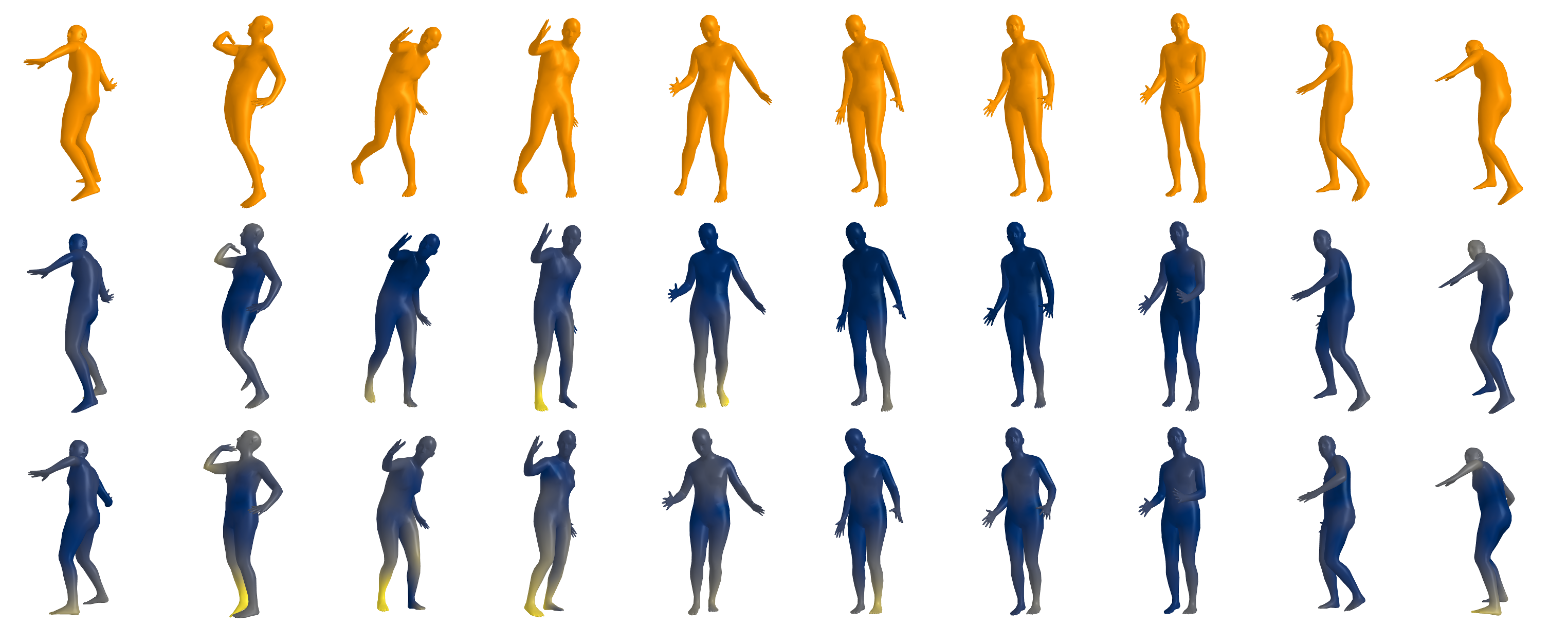}
    \caption{Qualitative comparison to the state of the part method~\cite{jiang2022avatarposer} in MC scenario. \textbf{Top}: Ground truth in orange, \textbf{Middle}: HMD-NeMo, \textbf{Bottom}: Jiang et al.~\cite{jiang2022avatarposer}.}
    \label{fig:supp_vis9}
\end{figure*}

\begin{figure*}
    \centering
    \includegraphics[width=\textwidth]{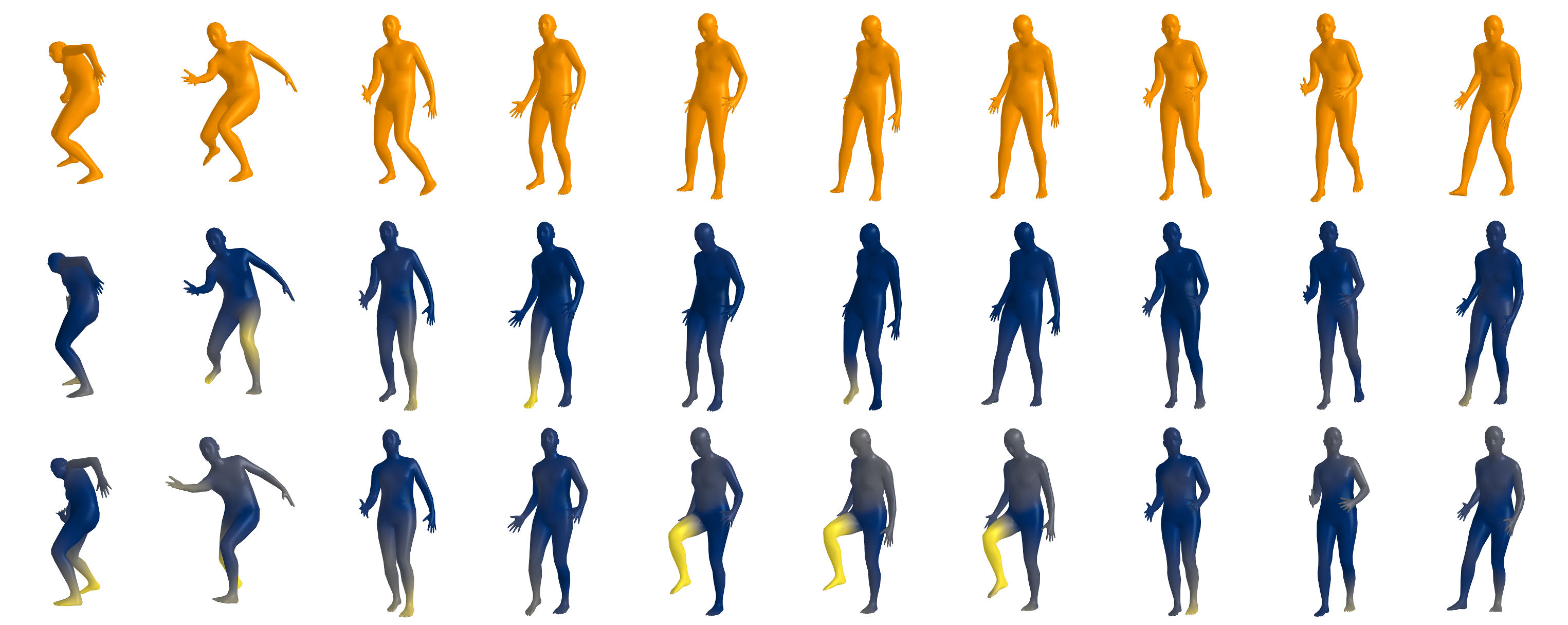}
    \caption{Qualitative comparison to the state of the part method~\cite{jiang2022avatarposer} in MC scenario. \textbf{Top}: Ground truth in orange, \textbf{Middle}: HMD-NeMo, \textbf{Bottom}: Jiang et al.~\cite{jiang2022avatarposer}.}
    \label{fig:supp_vis10}
\end{figure*}

\begin{figure*}
    \centering
    \includegraphics[width=\textwidth]{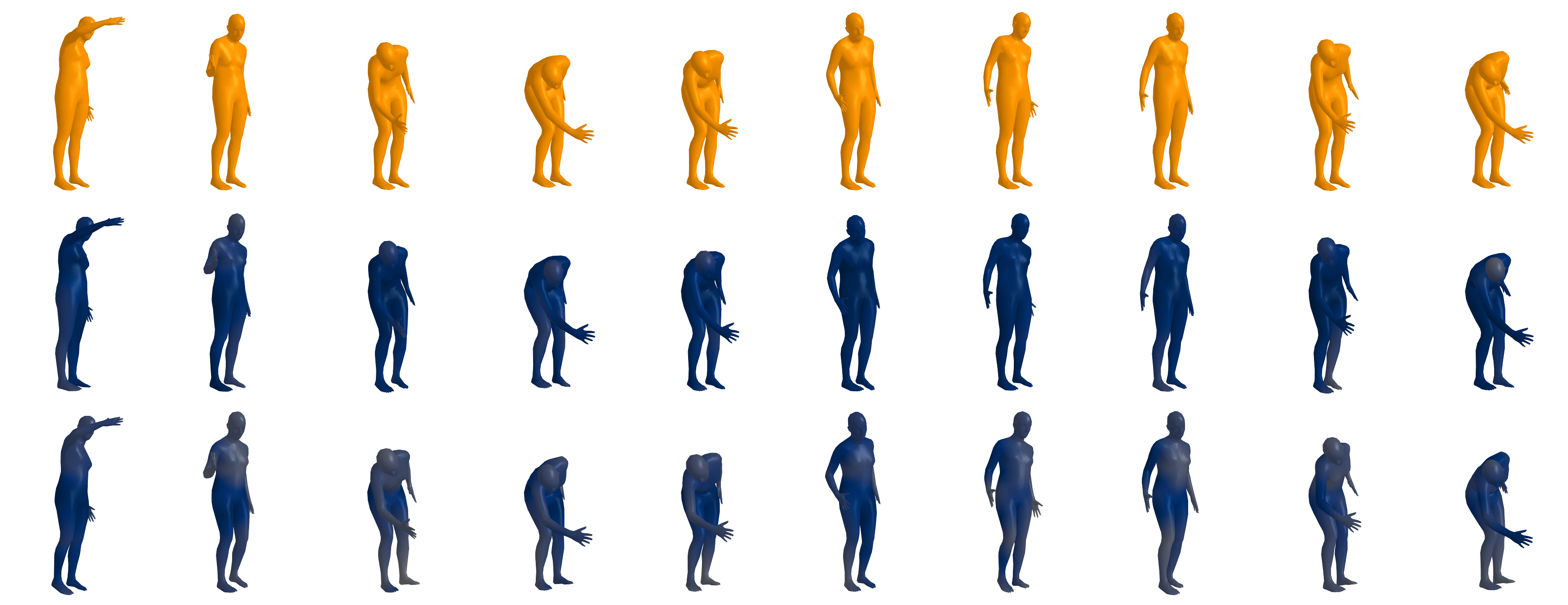}
    \caption{Qualitative comparison to the state of the part method~\cite{jiang2022avatarposer} in MC scenario. \textbf{Top}: Ground truth in orange, \textbf{Middle}: HMD-NeMo, \textbf{Bottom}: Jiang et al.~\cite{jiang2022avatarposer}.}
    \label{fig:supp_vis11}
\end{figure*}

\begin{figure*}
    \centering
    \includegraphics[width=\textwidth]{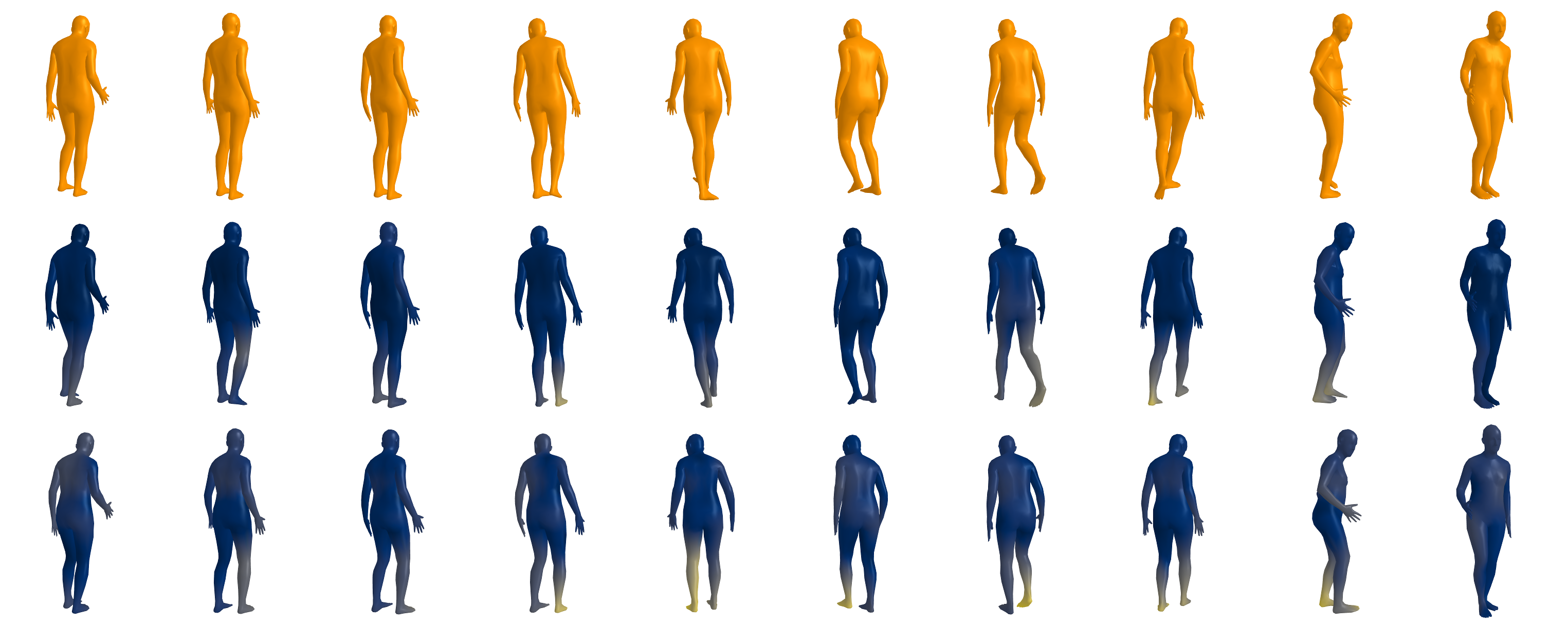}
    \caption{Qualitative comparison to the state of the part method~\cite{jiang2022avatarposer} in MC scenario. \textbf{Top}: Ground truth in orange, \textbf{Middle}: HMD-NeMo, \textbf{Bottom}: Jiang et al.~\cite{jiang2022avatarposer}.}
    \label{fig:supp_vis12}
\end{figure*}

\begin{figure*}
    \centering
    \includegraphics[width=\textwidth]{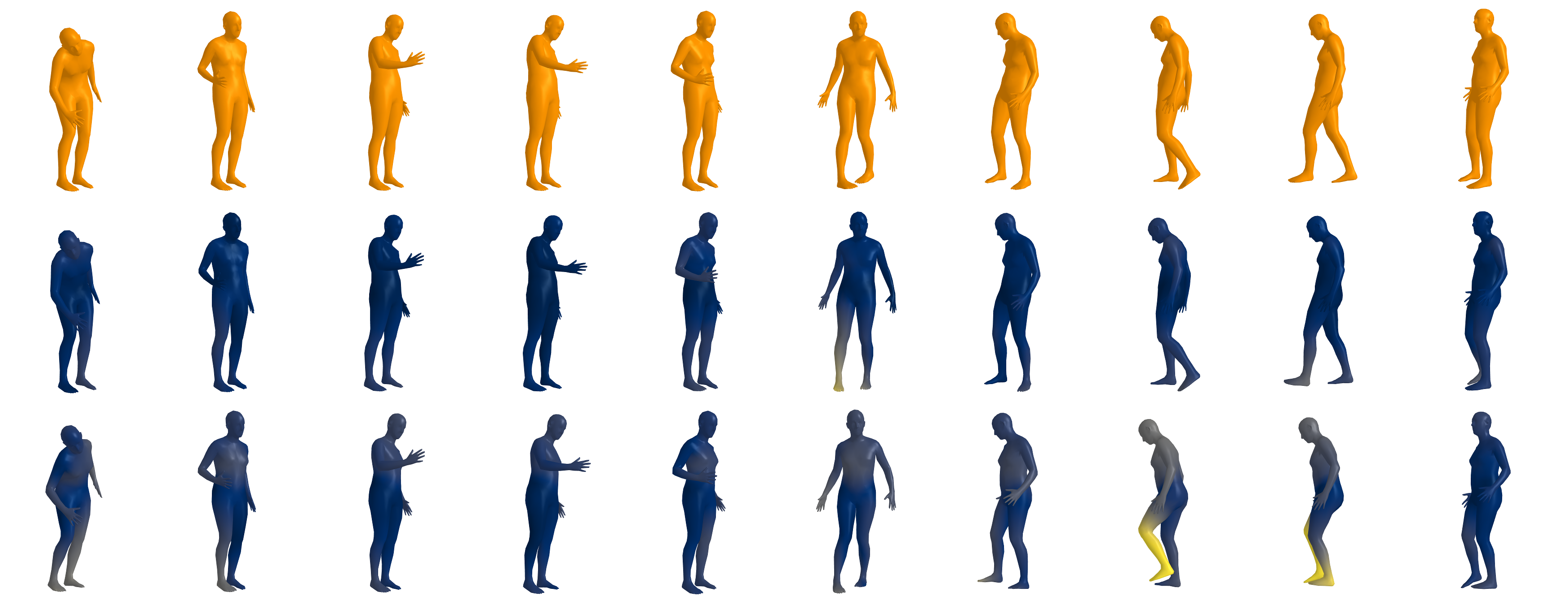}
    \caption{Qualitative comparison to the state of the part method~\cite{jiang2022avatarposer} in MC scenario. \textbf{Top}: Ground truth in orange, \textbf{Middle}: HMD-NeMo, \textbf{Bottom}: Jiang et al.~\cite{jiang2022avatarposer}.}
    \label{fig:supp_vis13}
\end{figure*}

\begin{figure*}
    \centering
    \includegraphics[width=\textwidth]{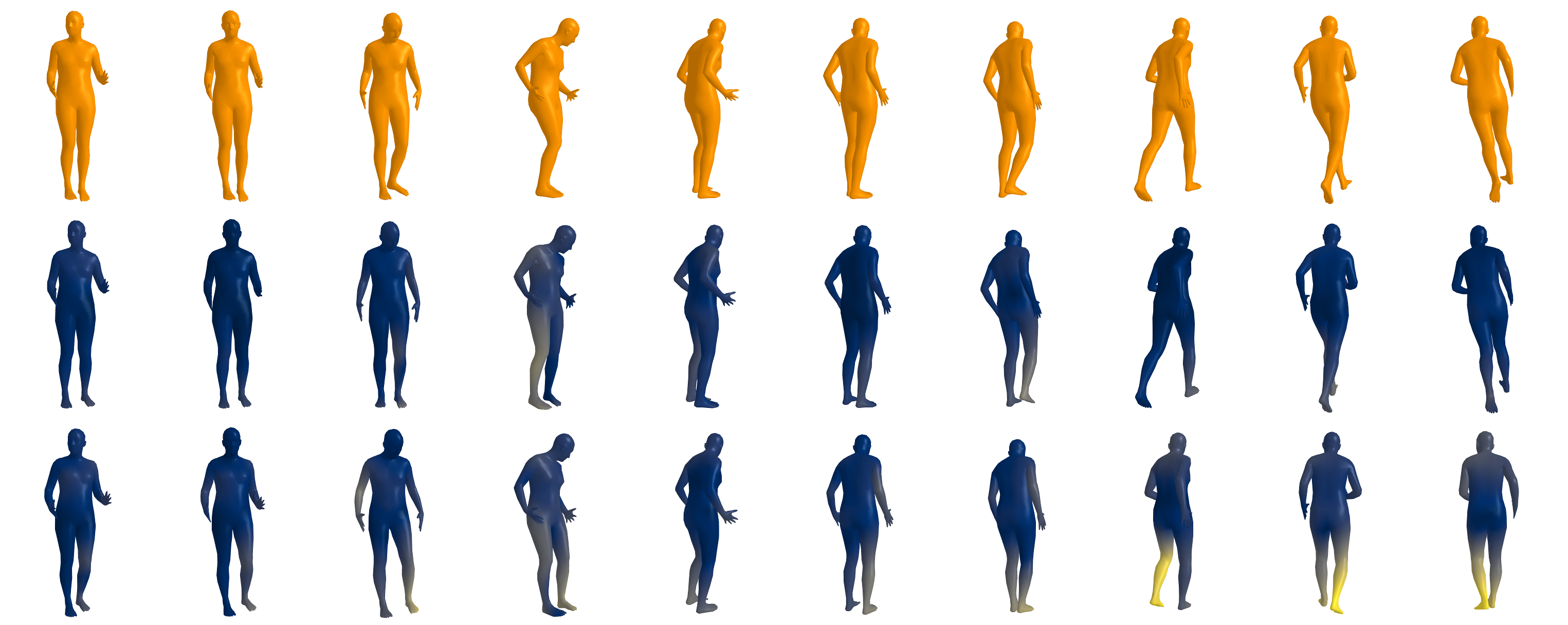}
    \caption{Qualitative comparison to the state of the part method~\cite{jiang2022avatarposer} in MC scenario. \textbf{Top}: Ground truth in orange, \textbf{Middle}: HMD-NeMo, \textbf{Bottom}: Jiang et al.~\cite{jiang2022avatarposer}.}
    \label{fig:supp_vis14}
\end{figure*}

\begin{figure*}
    \centering
    \includegraphics[width=\textwidth]{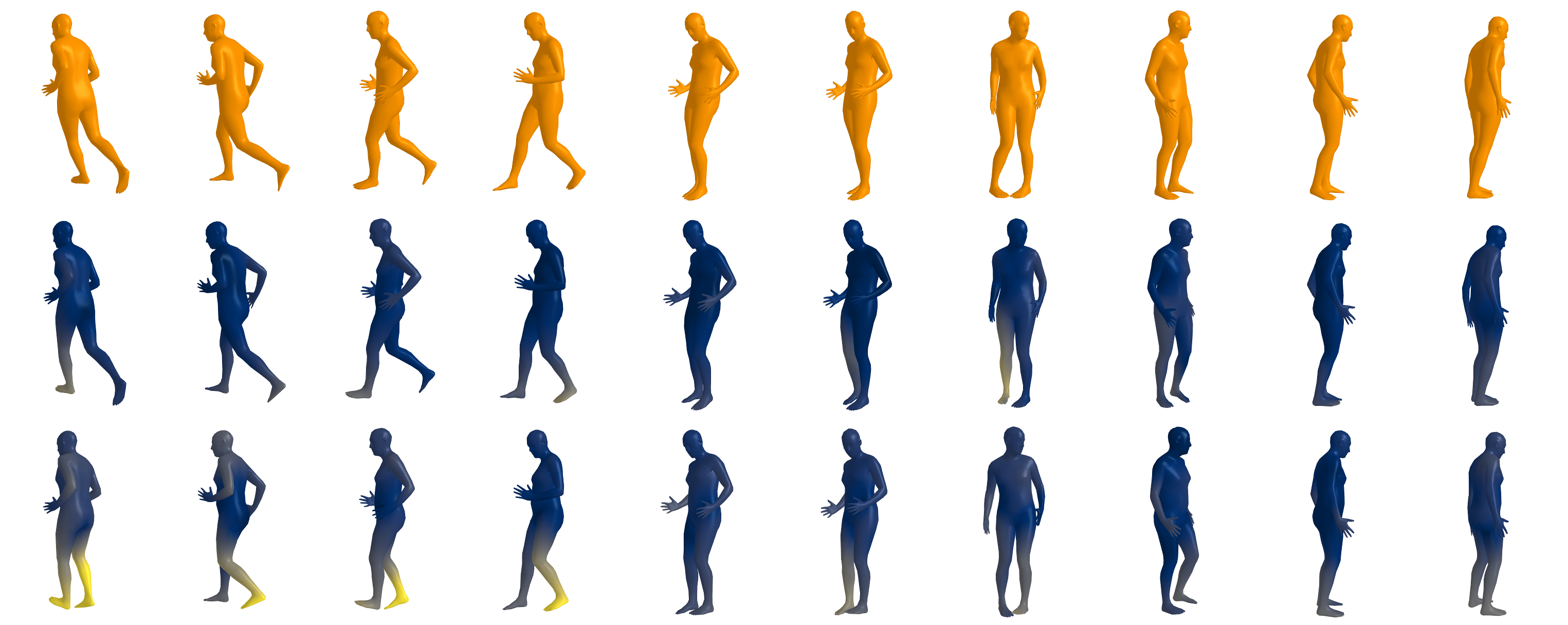}
    \caption{Qualitative comparison to the state of the part method~\cite{jiang2022avatarposer} in MC scenario. \textbf{Top}: Ground truth in orange, \textbf{Middle}: HMD-NeMo, \textbf{Bottom}: Jiang et al.~\cite{jiang2022avatarposer}.}
    \label{fig:supp_vis15}
\end{figure*}

\begin{figure*}
    \centering
    \includegraphics[width=\textwidth]{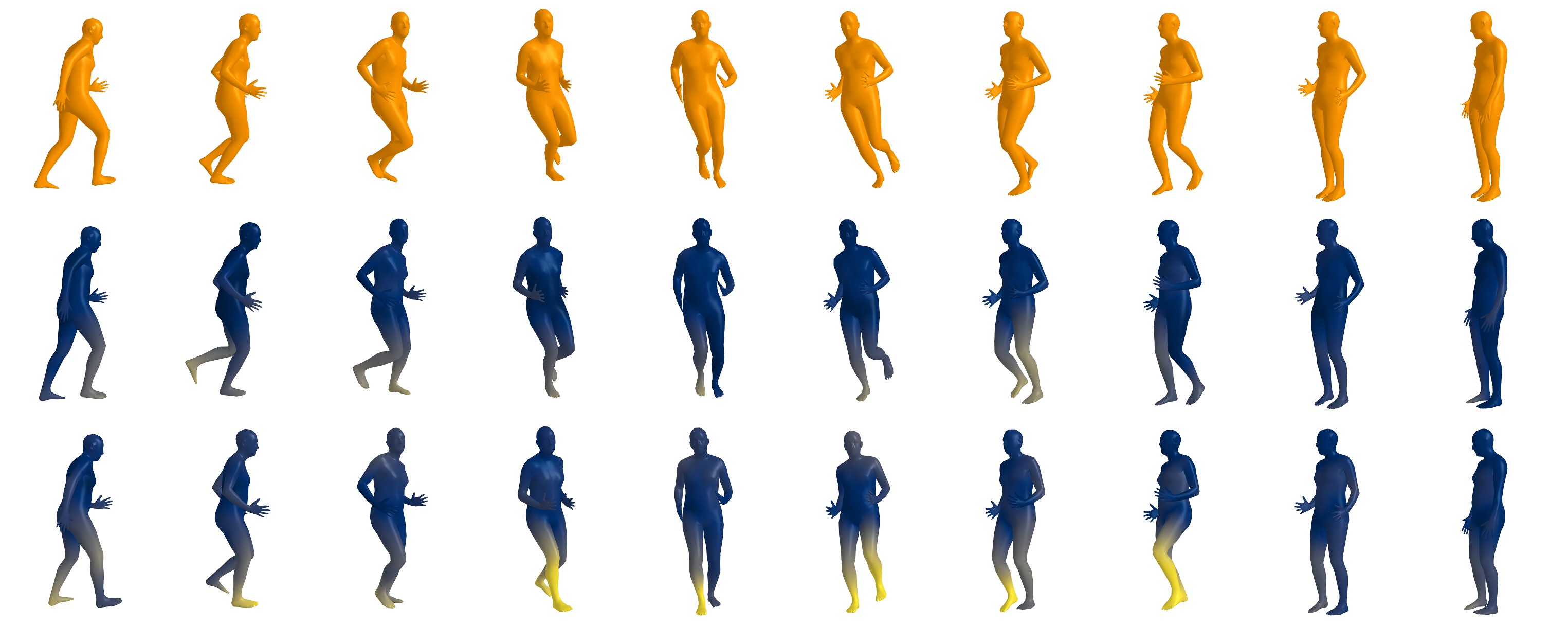}
    \caption{Qualitative comparison to the state of the part method~\cite{jiang2022avatarposer} in MC scenario. \textbf{Top}: Ground truth in orange, \textbf{Middle}: HMD-NeMo, \textbf{Bottom}: Jiang et al.~\cite{jiang2022avatarposer}.}
    \label{fig:supp_vis16}
\end{figure*}

\begin{figure*}
    \centering
    \includegraphics[width=\textwidth]{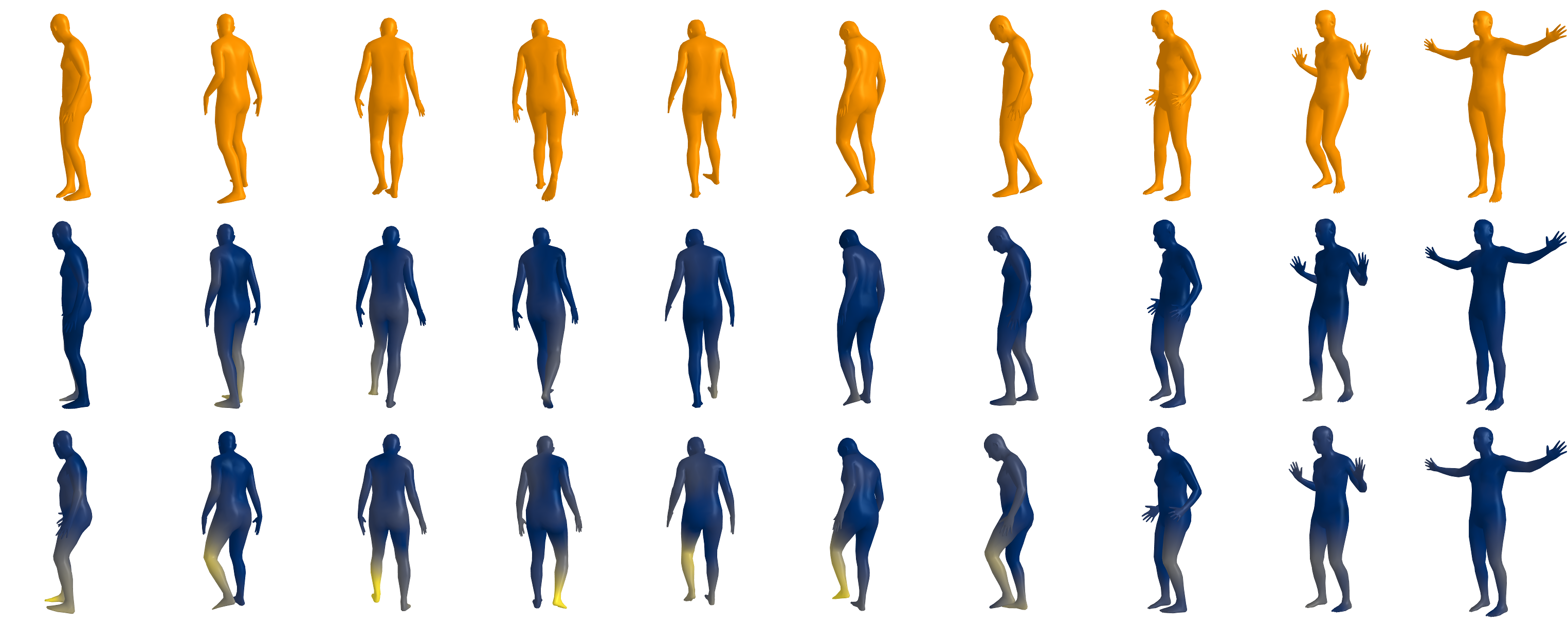}
    \caption{Qualitative comparison to the state of the part method~\cite{jiang2022avatarposer} in MC scenario. \textbf{Top}: Ground truth in orange, \textbf{Middle}: HMD-NeMo, \textbf{Bottom}: Jiang et al.~\cite{jiang2022avatarposer}.}
    \label{fig:supp_vis17}
\end{figure*}

\begin{figure*}
    \centering
    \includegraphics[width=\textwidth]{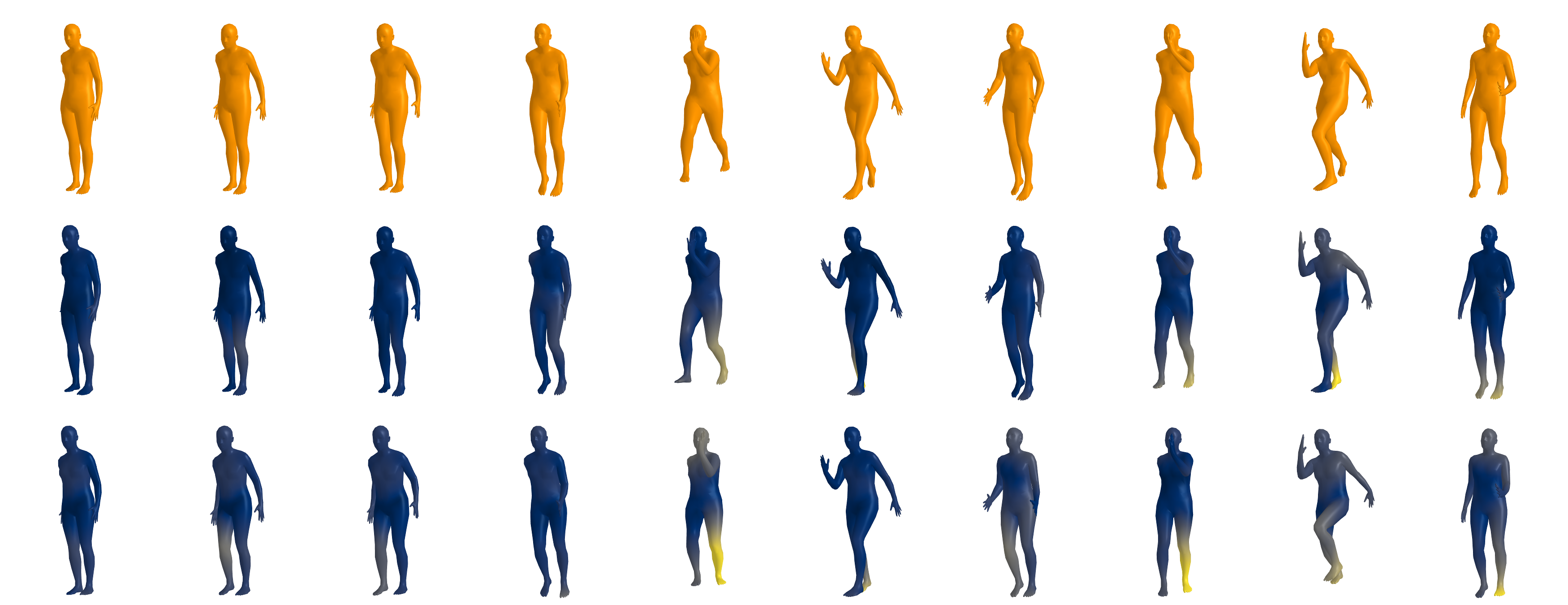}
    \caption{Qualitative comparison to the state of the part method~\cite{jiang2022avatarposer} in MC scenario. \textbf{Top}: Ground truth in orange, \textbf{Middle}: HMD-NeMo, \textbf{Bottom}: Jiang et al.~\cite{jiang2022avatarposer}.}
    \label{fig:supp_vis18}
\end{figure*}

\begin{figure*}
    \centering
    \includegraphics[width=\textwidth]{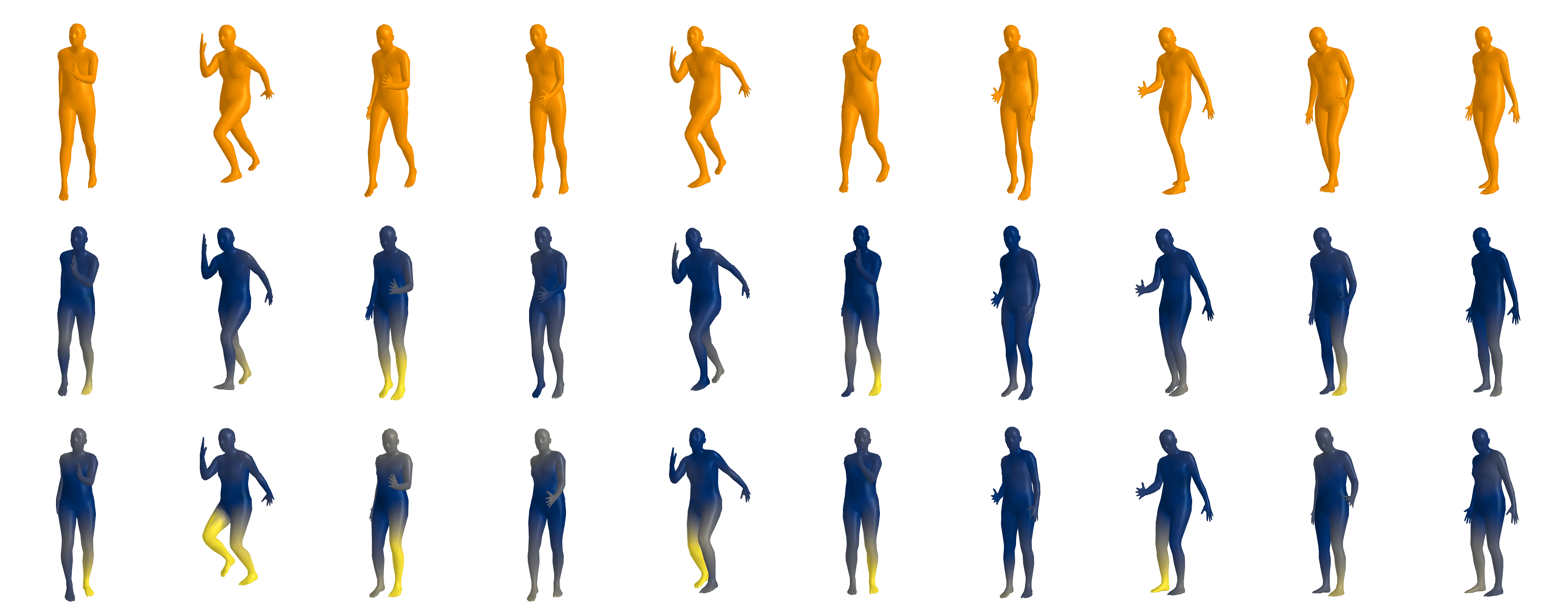}
    \caption{Qualitative comparison to the state of the part method~\cite{jiang2022avatarposer} in MC scenario. \textbf{Top}: Ground truth in orange, \textbf{Middle}: HMD-NeMo, \textbf{Bottom}: Jiang et al.~\cite{jiang2022avatarposer}.}
    \label{fig:supp_vis19}
\end{figure*}

\begin{figure*}
    \centering
    \includegraphics[width=\textwidth]{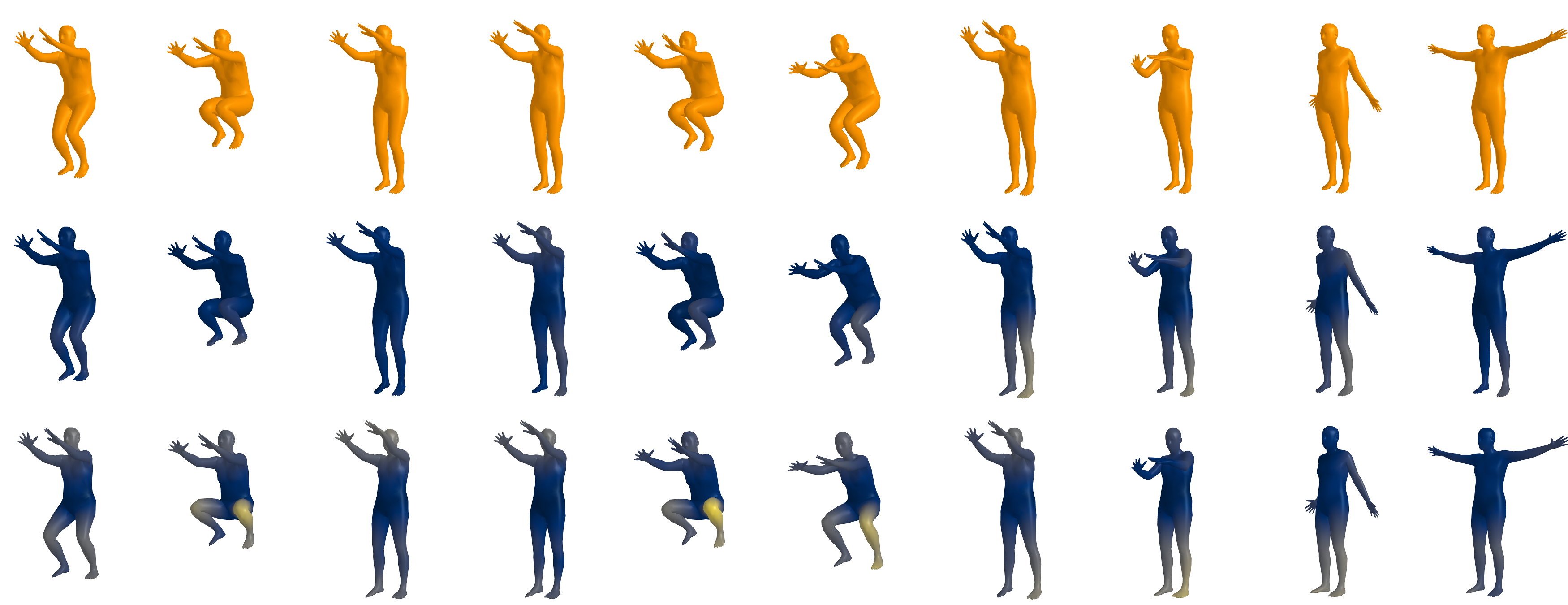}
    \caption{Qualitative comparison to the state of the part method~\cite{jiang2022avatarposer} in MC scenario. \textbf{Top}: Ground truth in orange, \textbf{Middle}: HMD-NeMo, \textbf{Bottom}: Jiang et al.~\cite{jiang2022avatarposer}.}
    \label{fig:supp_vis20}
\end{figure*}

\begin{figure*}
    \centering
    \includegraphics[width=\textwidth]{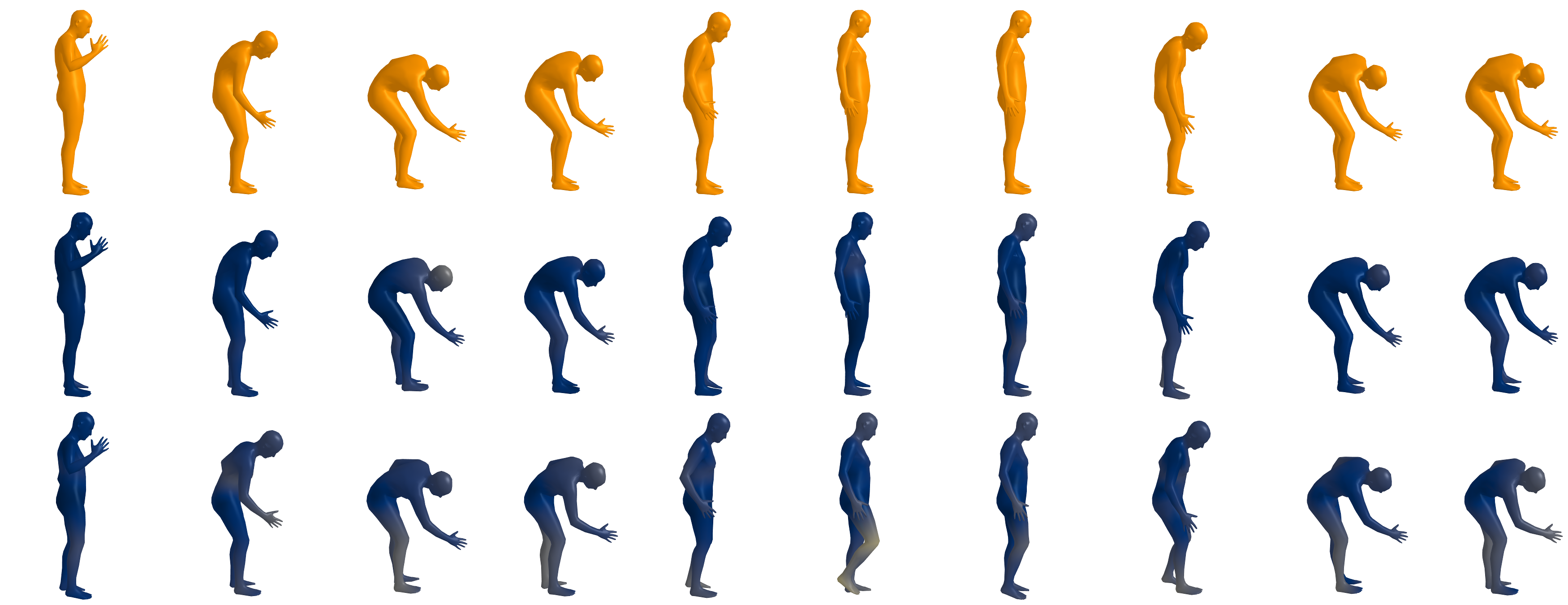}
    \caption{Qualitative comparison to the state of the part method~\cite{jiang2022avatarposer} in MC scenario. \textbf{Top}: Ground truth in orange, \textbf{Middle}: HMD-NeMo, \textbf{Bottom}: Jiang et al.~\cite{jiang2022avatarposer}.}
    \label{fig:supp_vis21}
\end{figure*}

\begin{figure*}
    \centering
    \includegraphics[width=\textwidth]{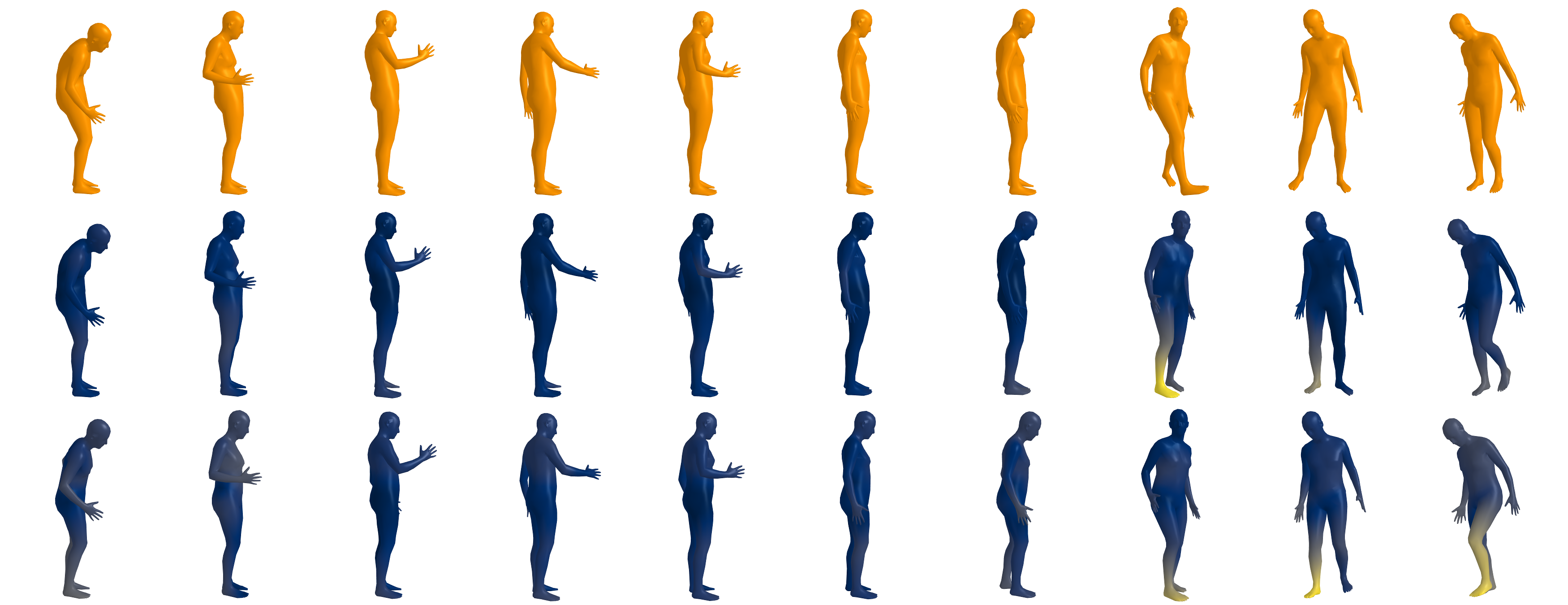}
    \caption{Qualitative comparison to the state of the part method~\cite{jiang2022avatarposer} in MC scenario. \textbf{Top}: Ground truth in orange, \textbf{Middle}: HMD-NeMo, \textbf{Bottom}: Jiang et al.~\cite{jiang2022avatarposer}.}
    \label{fig:supp_vis22}
\end{figure*}

\begin{figure*}
    \centering
    \includegraphics[width=\textwidth]{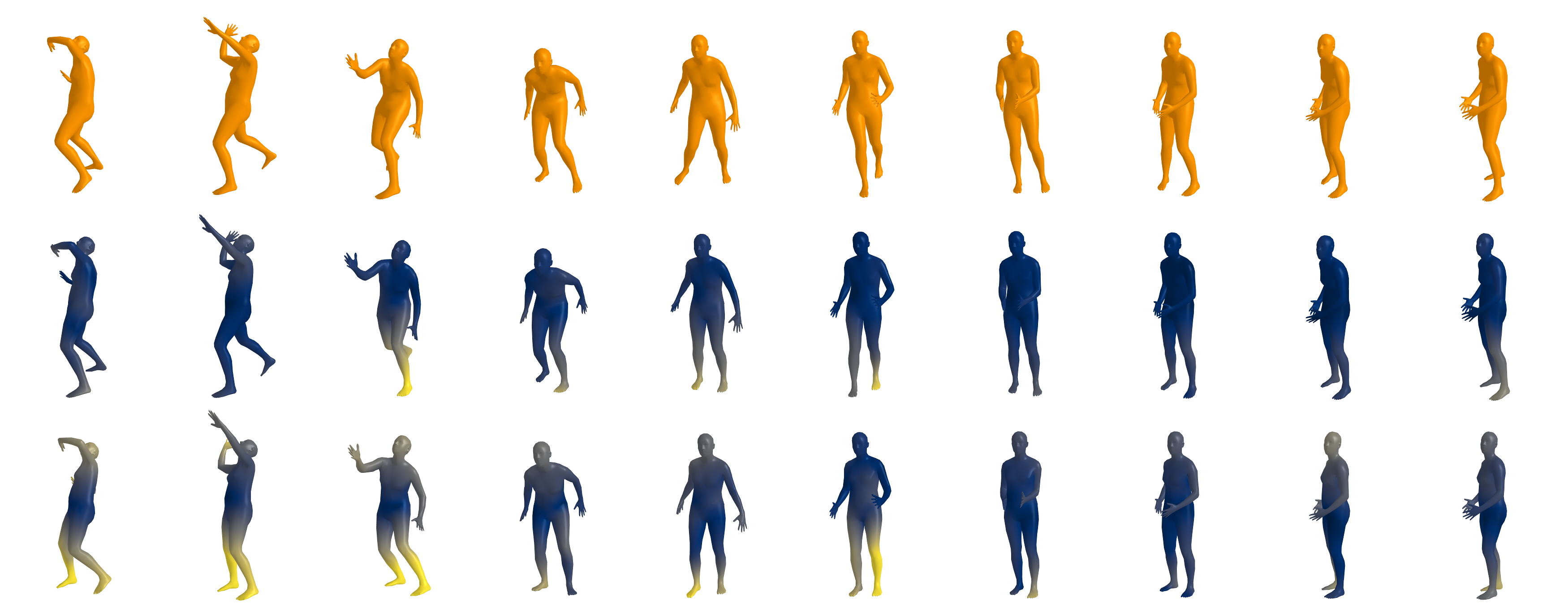}
    \caption{Qualitative comparison to the state of the part method~\cite{jiang2022avatarposer} in MC scenario. \textbf{Top}: Ground truth in orange, \textbf{Middle}: HMD-NeMo, \textbf{Bottom}: Jiang et al.~\cite{jiang2022avatarposer}.}
    \label{fig:supp_vis23}
\end{figure*}

\begin{figure*}
    \centering
    \includegraphics[width=\textwidth]{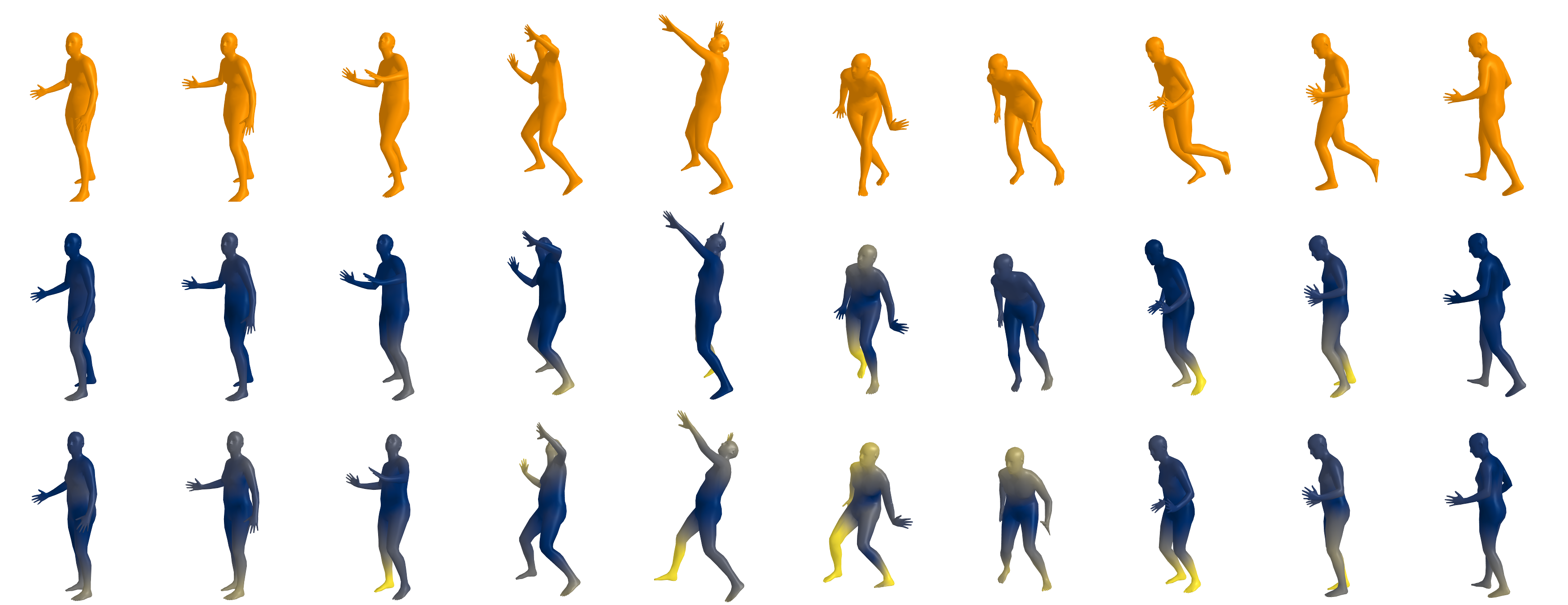}
    \caption{Qualitative comparison to the state of the part method~\cite{jiang2022avatarposer} in MC scenario. \textbf{Top}: Ground truth in orange, \textbf{Middle}: HMD-NeMo, \textbf{Bottom}: Jiang et al.~\cite{jiang2022avatarposer}.}
    \label{fig:supp_vis24}
\end{figure*}

\begin{figure*}
    \centering
    \includegraphics[width=\textwidth]{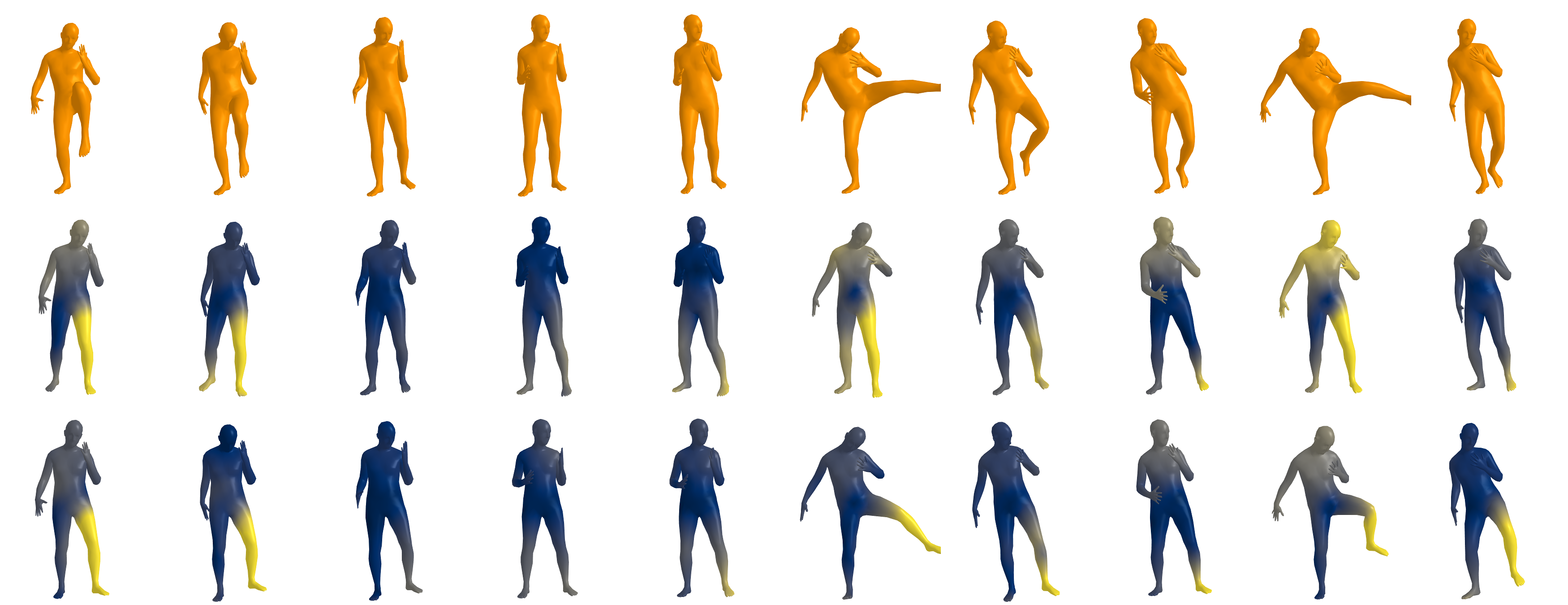}
    \caption{Qualitative comparison to the state of the part method~\cite{jiang2022avatarposer} in MC scenario. \textbf{Top}: Ground truth in orange, \textbf{Middle}: HMD-NeMo, \textbf{Bottom}: Jiang et al.~\cite{jiang2022avatarposer}.}
    \label{fig:supp_vis25}
\end{figure*}

\begin{figure*}
    \centering
    \includegraphics[width=\textwidth]{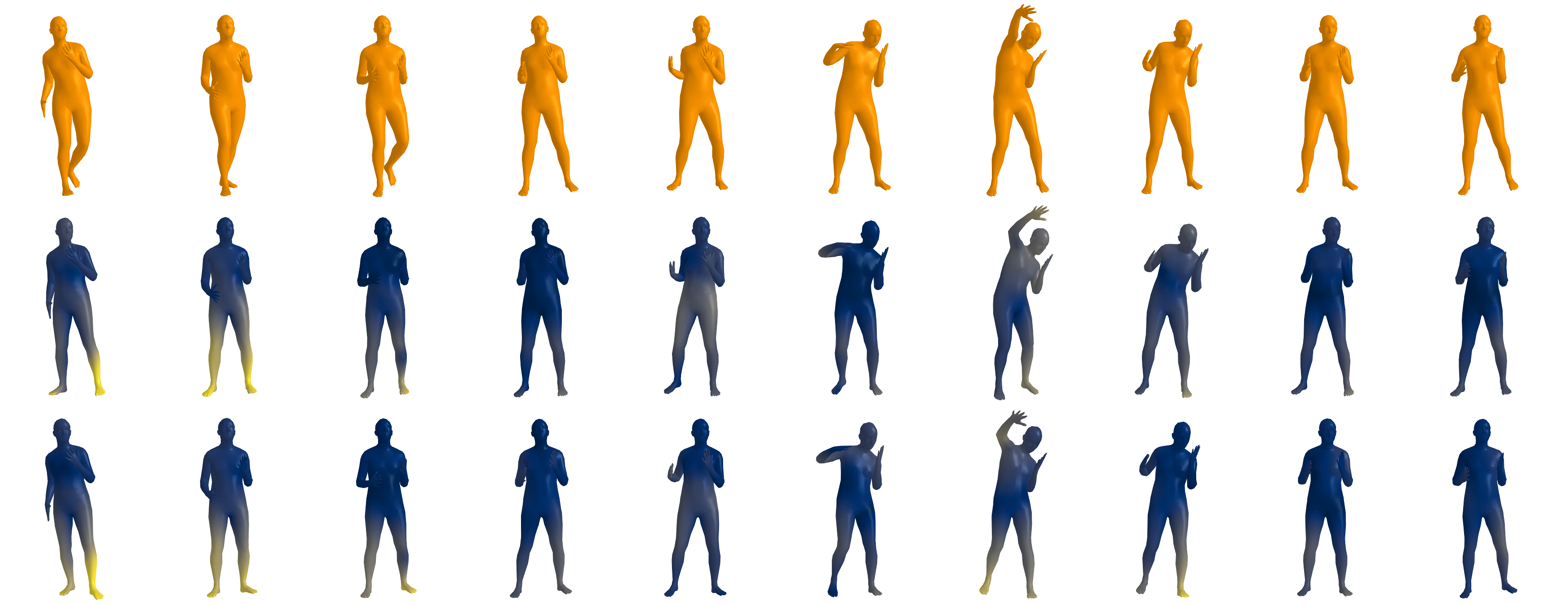}
    \caption{Qualitative comparison to the state of the part method~\cite{jiang2022avatarposer} in MC scenario. \textbf{Top}: Ground truth in orange, \textbf{Middle}: HMD-NeMo, \textbf{Bottom}: Jiang et al.~\cite{jiang2022avatarposer}.}
    \label{fig:supp_vis26}
\end{figure*}

In this section, we first provider results of HMD-NeMo (see Fig.~\ref{fig:supp_vis1} to Fig.~\ref{fig:supp_vis4}) and then provide more qualitative comparisons to the state-of-the-art approach~\cite{jiang2022avatarposer} (see Fig.~\ref{fig:supp_vis5} to Fig.~\ref{fig:supp_vis26}). Note that none of the results are cherry picked. Note that all results are color-coded based on the distance between the predicted vertices and that of the ground truth. Dark blue vertices represent predictions that are very close to the ground truth and yellow vertices are further away from ground truth. 

In comparison to~\cite{jiang2022avatarposer}, both our approach and the baseline performs reasonably well in predicting the upper-body mainly due to the fact that HMD signal is a strong signal about the upper body pose. Typically, HMD-NeMo does a relatively better job at predicting more plausible lower body motion. 

\end{document}